\documentclass[journal]{IEEEtran}
\usepackage{amsmath}
\usepackage[colorlinks,linkcolor=blue]{hyperref}
\usepackage{graphicx}
\usepackage{amssymb}
\usepackage{booktabs}
\usepackage{amssymb}
\usepackage{indentfirst}
\usepackage[misc]{ifsym}
\usepackage{amsmath}
\usepackage{multirow}
\usepackage{booktabs}
\usepackage{tikz,xcolor,hyperref}
\usepackage{bbding}
\usepackage{underscore}
\usepackage{amssymb}
\usepackage{amssymb}
\makeatletter

\newcommand{\Rmnum}[1]{\expandafter\@slowromancap\romannumeral #1@}
\makeatother

\ifCLASSINFOpdf
\else
\fi
\begin{document}
%
\title{SFANet: A Spectrum-aware Feature Augmentation Network for Visible-Infrared Person Re-Identification}
%
%
%

\author{Haojie Liu,
        Shun Ma,
        Daoxun Xia,
        and Shaozi Li,~\IEEEmembership{Senior Member, IEEE}
\thanks{\textit{Corresponding author: Daoxun Xia, e-mail: dxxia@gznu.edu.cn}}
\thanks{H. Liu and D. Xia are with the School of Big Data and Computer Science and also with Engineering Laboratory for Applied Technology of Big Data in Education, Guizhou Normal University, Guiyang 550025, China (e-mail: 19010230505@gznu.edu.cn, dxxia@gznu.edu.cn).}
\thanks{S. Ma is with the School of Big Data and Computer Science, Guizhou Normal University, Guiyang 550025, China (e-mail: 870796594@qq.com).}
\thanks{S. Li is with School of Informatics, Xiamen University, Xiamen 361005, China (e-mail: szlig@xmu.edu.cn).}}

%
%

\markboth{Journal of \LaTeX\ Class Files,~Vol.~14, No.~8, August~2015}%
{Shell \MakeLowercase{\textit{et al.}}: Bare Demo of IEEEtran.cls for IEEE Journals}
%



\maketitle

\begin{abstract} Visible-Infrared person re-identification (VI-ReID) is a challenging matching problem due to large modality varitions between visible and infrared images. Existing approaches usually bridge the modality gap with only feature-level constraints, ignoring pixel-level variations. Some methods employ GAN to generate style-consistent images, but it destroys the structure information and incurs a considerable level of noise. In this paper, we explicitly consider these challenges and formulate a novel spectrum-aware feature augementation network named SFANet for cross-modality matching problem. Specifically, we put forward to employ grayscale-spectrum images to fully replace RGB images for feature learning. Learning with the grayscale-spectrum images, our model can apparently reduce modality discrepancy and detect inner structure relations across the different modalities, making it robust to color variations. In feature-level, we improve the conventional two-stream network through balancing the number of specific and sharable convolutional blocks, which preserve the spatial structure information of features. Additionally, a bi-directional tri-constrained top-push ranking loss (BTTR) is embedded in the proposed network to improve the discriminability, which efficiently further boosts the matching accuracy. Meanwhile, we further introduce an effective dual-linear with batch normalization ID embedding method to model the identity-specific information and assits BTTR loss in magnitude stabilizing. On SYSU-MM01 and RegDB datasets, we conducted extensively experiments to demonstrate that our proposed framework contributes indispensably and achieves a very competitive VI-ReID performance.

\end{abstract}


%
\IEEEpeerreviewmaketitle
\section{Introduction}
\IEEEPARstart{P}{erson} Re-identification (Re-ID) aims at identifying the same person across multiple non-overlapping camera views \cite{1, 2}. It has been widely investigated because of its great importance for video surveilance \cite{3, 4}. Existing Re-ID methods mainly focus on RGB-RGB matching problem \cite{7, 8}, where all images are captured under good visible light conditions. However, in low-illumination environment, visible cameras cannot provide enough discriminative characteristics of a person, as shown in Fig. 1 (a). Against this issue, \textit{Visible-Infrared Person Re-Identification} (VI-ReID) is introduced for person matching, which is imperative for practical surveillance applications.

As shown in Fig. 1 (b), given a target infrared (RGB) image, the main goal of VI-ReID is to search the corresponding RGB (infrared) images from a gallery set captured by other spectrum cameras \cite{13, 5, 11, 12, 6}. To our best knowledge, very few works have paid attention to the visible-infrared person re-identification mainly for two reasons. First, VI-ReID suffers from the large modality discrepancy between imaging processes of different spectrum cameras. Second, the person's appearance discrepancy caused by distinct viewpoints, pose variations and scale changes give in large intra-class variations. As a result, the neural network is difficult to find such a shared feature space that the different modality information can be treated equally.
\begin{figure}[tp]
	\begin{center}
		\includegraphics[width=0.99\linewidth]{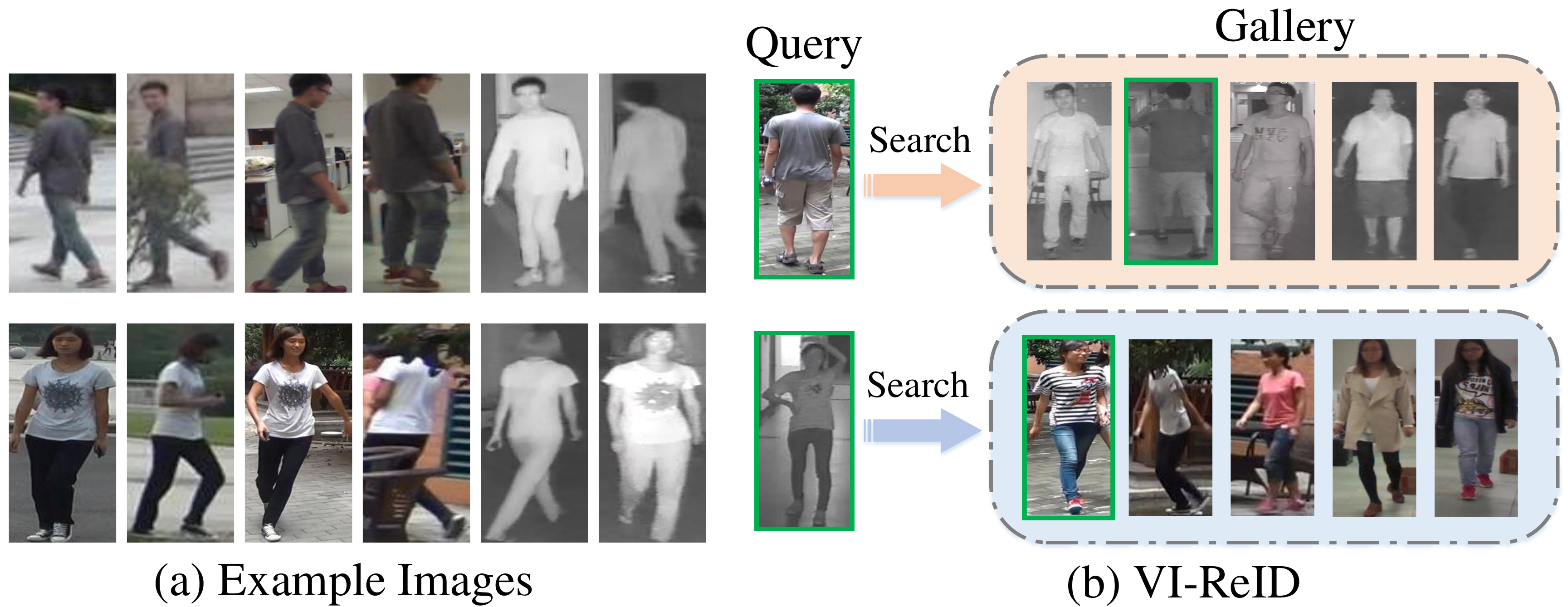}\vspace{-0.4cm}
	\end{center}
	\caption{(a) Example images from the SYSU-MM01 dataset [5]. (b) Illustration of visible-infrared cross-modality person re-identification. Given a visible image, the image of the corresponding identity person should be found from the infrared gallery, and vice versa. \vspace{-0.6cm}}
	\label{fig:smalltarget}
\end{figure}

To deal with the above issues, some pioneer works have been proposed. Wu et al. \cite{5} introduced a deep zero-padding method to extract unified cross-modality feature vectors by adaptively handling the modality input. Contemporarily, Ye et al. proposed an effective two-stream framework in \cite{11} for feature learning and metric learning. \cite{12} improved this idea and introduced a dual-constrained top-ranking loss to learn discriminative feature representations. However, these methods only consider the modality discrepancy of global coarse-grained features, but ignore the variation of pixel-level fine-grained features. More recently, some GAN-based methods \cite{10, 14, 16, 15} have been proposed to generate cross-modality images to fill the gap between two different modalities. However, GAN based model training is unstable and the key local structure information of images is easily destroyed with unsupervised generation manners.
\begin{figure}[tp]
	\begin{center}
		\includegraphics[width=0.99\linewidth]{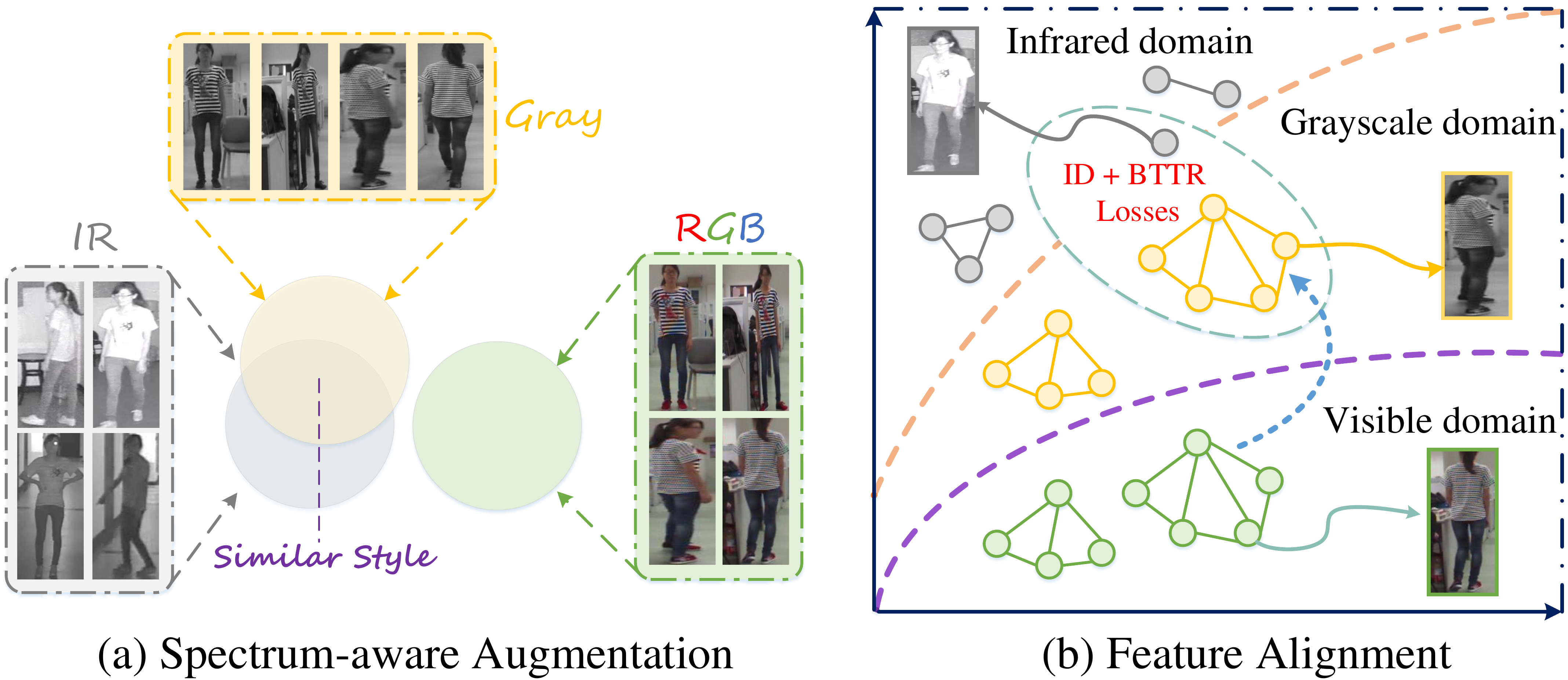}\vspace{-0.4cm}
	\end{center}
	\caption{Illustration of our solution for the challenging VI-ReID task. (a) Grayscale-spectrum images are treated as feature augmention strategy to approximate the infrared image style for reducing modality discrepancy. (b) The proposed "dual-linear with batch normalization ID embedding method" and the strong constraint "bi-directional tri-constrained top-push ranking loss" are used to guide the taining process.\vspace{-0.6cm}}
	\label{fig:smalltarget}
\end{figure}

In view of the analysis above, we formulate a novel spectrum-aware feature augementation network, capable of alleviating the large cross-modality discrepancy in image-level and providing strong feature-level constraints to derives the discriminative property to distinguish the images between different classes. Our main idea is illustrated in Fig. 2. Different from the approaches \cite{14, 16, 10} that use GAN to generate modality-consistent images, we carry out cross-spectrum feature augmentation by generating spectrum-aware grayscale images to transform the original RGB features into a new homogeneously augmented space. The generated grayscale images are effective for model training because they approximate the infrared style and meanwhile preserve complete structure information of visible images. In this situation, to further alleviate remaining cross-modality discrepancy between grayscale spectrum and infrared modality, we adopt a sharable dual-path information-preserving structure with sharing some parameters of residual blocks which can be treated as the 3D-shaped tensor spaces to perserve the image information of spatial structures. This parameter sharing strategy was introduced in \cite{47}, which can be set as a strong and effective baseline backbone.

Additionally, we also investigate cross-modality sharable identity-specific information and explore a satisfactory metric measure to minimize the ambiguity between classes while maximizing cross-modality similarity among instances, with the aim to make person features towards modality-invariant for facilitating the cross-modality adaptation process using SFANet. More concretely, we propose a strong bi-directional tri-constrained top-push ranking (BTTR) loss with the separate cross-modality, intra-modality and inter-modality training strategy to guide the training process. Compared with the BDTR loss introduced in \cite{12}, the designed novel inter-modality regularizer significantly considers the distance relationship between cross-modality and intra-modaltiy, addressing the difficulty in learning discriminative feature embedding. Furthermore, a dual-linear with batch normalization ID embedding method is employed to improve the robustness against modality varitions in classifier-level. The in-depth analysis of SFNet is presented in section ``Discussion". 


The main contributions can be summarized in four-fold:

$\bullet$ We propose a spectrum-aware feature augementation network for visible-infrared person Re-ID. To our best knowledge, this is the first attempt to expoilt grayscale-spetrum images to fully replace conventional RGB images for cross-modality feature learning. 

$\bullet$ We develop a sharable dual-path information-preserving network to share the parameters of two shallow convolutional layers for feature embedding. Compared with the first parameter sharing work for VI-ReID in \cite{33}, it achieves a more outstanding performance.

$\bullet$ We propose the Bi-directional Tri-constrained Top-Push Ranking (BTTR) Loss to constrain the relative distance of different classes from both the same modality and cross-modality, further promoting the performance of the network.  

$\bullet$ To improve classification accuracy and help stablize the magnitudes of embedding vectors, we design a novel classifier structure, named "Dual-Linear with Batch Normalization ID Embedding", which bring large performance improvement working with the proposed BTTR loss function.

\begin{figure*}[t]
	\begin{center}
		\includegraphics[width=0.99\linewidth]{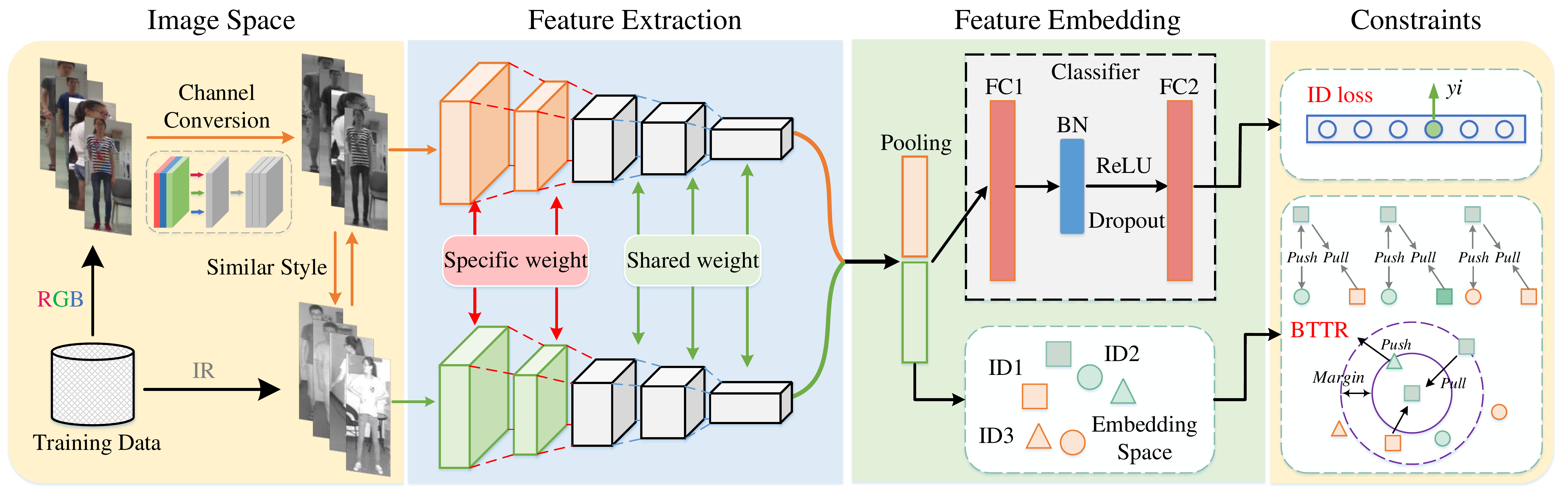}\vspace{-0.4cm}
	\end{center}
	\caption{The architecture of the proposed SFANet. We first exploit the channel conversion operation to generate cross-spectrum grayscale images. After a channel expansion, images from grayscale and infrared modalities are then fed into the sharable dual-path information-preserving network for feature learning. Two loss functions, \textit{i.e.}, cross-modality sharable identity loss and bi-directional tri-constrained top-push ranking loss, are used to supervise the training process.\vspace{-0.3cm}}
	\label{fig:smalltarget}
\end{figure*}

\section{Related Work}
In this section, we review the literature from the following three main research directions.

\textbf{Visible Camera-Based Re-ID Methods.} Person re-identification (Re-ID) mainly focus on the challenge of person's appearance changes such as different camera views, poses varitons, partial occlusion, and illuminations changes \cite{17, 18, 19, 20}. Traditional visible camera-based Re-ID methods can be divided into two fundamental components: feature extraction \cite{40, 41, 9, 21} and metric learning \cite{42, 43, 22, 39}. The former feature extraction works mainly focus on extracting more robust and discriminative feature representations. For example, Zhang et al. \cite{9} proposed a visual control flow language plus (VCFL+) model to improve the fusion process at the feature map level with the attention mechanism. The latter metric learning works aim at making a pair of true
matches have a relatively smaller distance than a pair of wrong matches. Liao et al. \cite{22} proposed a logic metric learning algorithm
which adopts an asymmetric sample weighting strategy to enhance Re-ID training. All these Re-ID models are effective enough in addressing the problem of appearance changes and have achieved human-level retrieval performance. However, they are only designed for single visible modality, which cannot perform well for cross-modality Re-ID task. 

\textbf{Cross-Modality Re-ID Methods.} Cross-modality person Re-ID aims at matching visible and infrared images of a person across different types, i.e. visible-to-infrared or infrared-to-visible \cite{5, 12, 6, 11, 23, 33}. Besides the appearance discrepancy in single-modality Re-ID, VI-ReID confronts a new challenge of large modality discrepancy between visible and infrared images. More related to our work, some great pioneer works have been made to explore the issuse of VI-ReID. In \cite{5}, Wu et al. collected the first cross-modality dataset named SYSU-MM01 and proposed a zero-padding strategy with one-stream network to adaptively evolve domain-specfic nodes for visible-infrared matching. Later on, Ye et al. \cite{11} proposed a hierarchical learning method to jointly optimize the modality-specific and modality-shared metrics. \cite{12} improved this idea and introduced a two-stream network with dual-constrained top-ranking loss to simultaneously handle the cross- and intra- modality variations. Besides, Dai et al. \cite{6} used generative adversarial training strategy to learn discriminative representations from different modalities. In addition, some methods \cite{23, 33} haved been proposed to improve the performance of feature learning by utilizing the modality-specific classifiers. However, all these methods always focus on filling the gap between visible and infrared inputs by means of feature-level constraints, ignoring the large cross-modality variation brought by different spectrum with dramatic pixel-level unbalance.

\textbf{Image generation for Person VI-ReID.} Recently developed generative adversarial technique provides a powerful tool for image translation. The work most relevant to ours are following five image-based methods \cite{25, 14, 15, 16, 10}. XIV cross-modality learning method was proposed in \cite{25}. It exploited a lightweight network to generate an auxiliary X modality with self-supervised manner for bridging modality gap between visible and infrared images. D$^2$RL \cite{14} used a generative sub-network to translate cross-modality images from given visible or infrared images and handled feature-level discrepancy through feature embedding in the unified space. AlignGAN \cite{10} is the first work to adopt the pixel-level and feature-level alignment strategies in a unified framework, which not only reduce the cross-modality and intra-modality discrepancy, but also learn identity-consitent features. Meanwhile, some other works \cite{15, 16} attempt to use GANs to generate more realistic cross-modality images to eliminate large modality discrenpancy. All these methods achieve superior performance, but training a great generator and a discriminator would cost huge computing resources and how to balance the generator and discriminator is an intractable problem. Comparatively, our generation with utilizing the linear accumulation of three RGB channels is easier to implement and the generated images have an identical underlying appearance with the original images.

\section{Our Approach} In this section, we first describe the feature augmentation of grayscale-spetrum images process, which addresses the large modality discrepancy with style simulation approach. Then, we introduce our sharable dual-path information-preserving network for discriminative feature extraction, which addresses the feature-level discrepancy with a partially shared structure. Finally, a novel batch normalized modality-shared classification and the homogeneous modality top-push ranking losses are introduced. The overall framework is shown in Fig. 3.

\subsection{Grayscale-spetrum Images Generation}
This subsection presents our grayscale-spetrum feature augmentation strategy which aims to approximate the style of single-channel infrared images to reduce the large modality discrepancy and retain the complete the feature information of person. Given an RGB image, we first extract each pixel value from the corresponding color channel (red, green, blue), and calculate the grayscale value with a transform function. Then, the grayscale pixel value is assigned to the corresponding position of the grayscale modality image, and all the pixels are traversed once to complete the transformation. 

For an RGB image $\mathcal{X}$ with red channel $R$, green channel $G$ and blue channel $B$, and a cross-spectrum image generation function $f$, the method which aims to generate corresponding grayscale image $\mathcal{Y}$ can be denoted as follows:
\begin{equation} 
\mathcal{X}(R,G,B)\stackrel{f}{\longrightarrow}\mathcal{Y},
\end{equation} 
\begin{equation} 
f(x)= \alpha R(x) + \beta G(x) + \delta B(x).
\end{equation}

Where the values of $\alpha$, $\beta$, and $\delta$ are 0.299, 0.587 and 0.114, respectively. Some examples are shown in Fig. 4. We can observe that the generated cross-spectrum images have similiar appearances to those captured by an infrared camera. In addition, during the generation process, we do not introduce any agnostic nosie that destroys the initial semantic information from RGB images. It denotes that our generated method can well approximate the image style of infrared images and maintain the structure information to the greatest extent.
\begin{figure}[t]
	\begin{center}
		\includegraphics[width=0.99\linewidth]{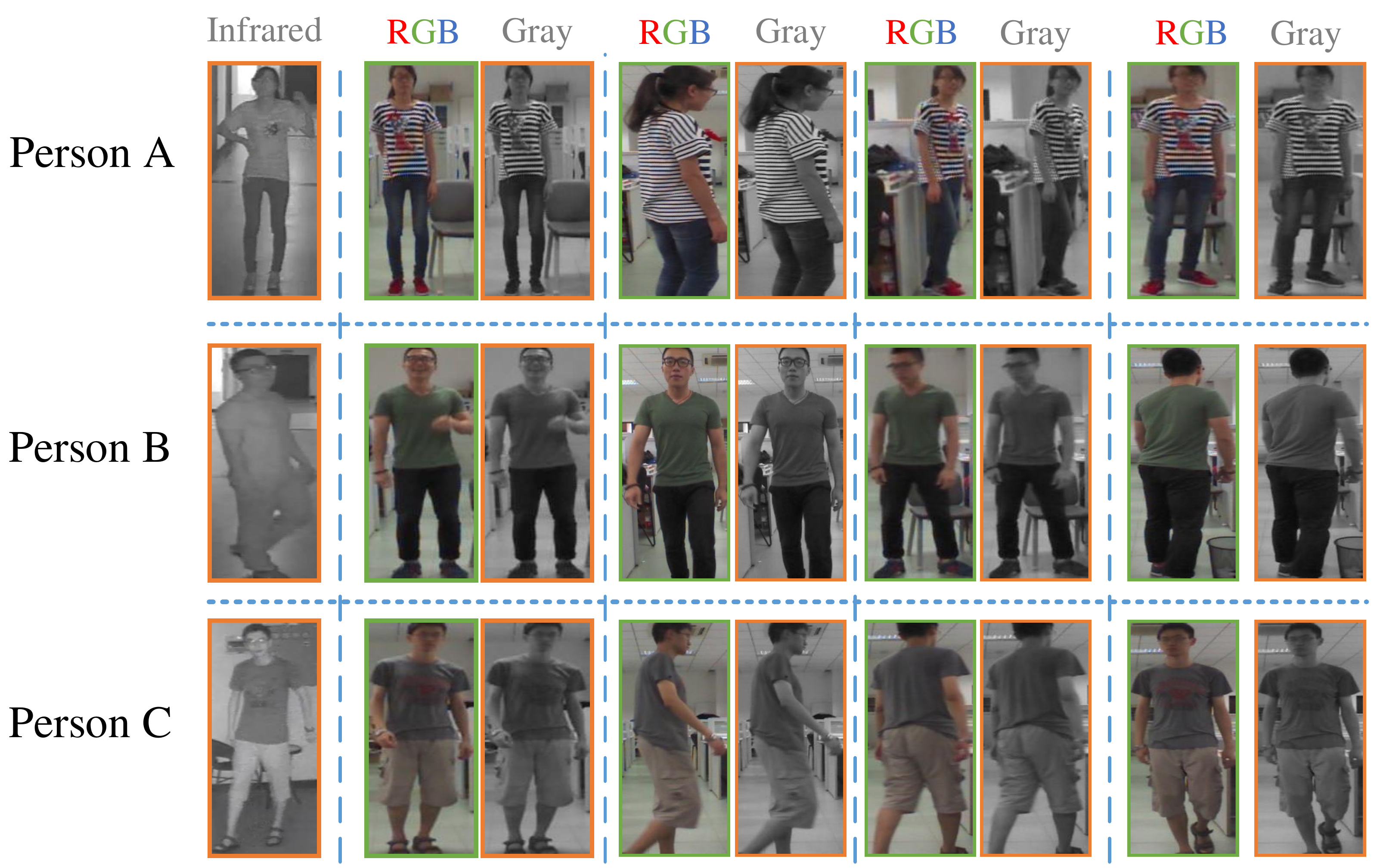}\vspace{-0.4cm}
	\end{center}
	\caption{Comparsion of our generated graycale images with RGB and IR images. Each row represents the same identity from SYSU-MM01 dataset. It can be observed that the images of the gray column are similar reference as the images captured by an infrared camera. \vspace{-0.3cm}}
	\label{fig:smalltarget}
\end{figure}

In this work, we utilize the grayscale images to replace visible images for model taining. Given visible inputs, our generation method can create point-to-point transformation grayscale images. However, these generated images have only single channel, while infrared images have three color channels. Therefore, we employ a channel expansion strategy to extend the single-channel images to three channels via simple copy operation, so all input images will have three channels that can be fed into a deep model.

\subsection{Sharable Dual-path Information-preserving Network.}
The two-stream network is a conventional way to extract features in VI-ReID task. However, the learned feature maps from this fully two-stream network are arbitrarily inconsistent due to distinct data distribution, which focus more on alleviating data gap between cameras while sacrificing discriminativeness, ignoring inherently correlation of the same person images from different cameras. Recently, parameter-sharing problem \cite{47} have been explored to analyze the impact of the number of parameter sharing for cross-modality feature learning. In this paper, we use the similar idea to refine both modality-sharable and modality-specific cues with parameter sharing strategy and introduce a sharable dual-path information-preserving network to extract discriminative common features of input images. The network contains two paths: grayscale-path and infrared path and both the paths are designed to share similar network structures in our cross-modality person re-identification task. Note that it mainly consists of two steps: modality specific feature extraction and modality shared feature embedding. The former focuses on capturing modality-specific low-level feature patterns, and the latter aims to learn common features of grayscale and infrared modalities.

As shown in Fig. 3, the generated three-channel grayscale images and infrared images are fed into the sharable dual-path information-preserving network. We use ResNet50 as our backbone network to extract feature maps of input images. The shallow two convolutional layers (Res-layer 0 and Res-layer 1) without sharing parameters are designed as feature extraction part to capture modality-specific low-level feature patterns. After that, the network parameters of convolution architectures on top of feature extraction part are shared to project the modality-specific inputs into a common space to learn modality-sharable high-level feature representations.

It is worth mentioning that compared with two-stream network proposed in \cite{12}, which utilizes the pre-trained five convolutional layers as independent feature extractor for fine-tuning, our improvement mainly lies in the partially shared network structures. This design enjoys processing 3D-shaped feature vectors instead of 1D-shaped feature maps computed by the shared fully-connected layer which losses rich person spatial structure information. To clarify, given an input tensor $\textit{x}_i^k$, the proposed network outputs a 3D common feature map $\textit{F}_i^k \in \mathbb{R}^{C\times H \times W}$, denoted by:
\begin{equation} 
\textit{F}_i^v=\mathcal{F}(\textit{\textbf{x}}_i^k; \Phi_{\mathcal{F}}), \forall k \in \left\{ v, t \right\},
\end{equation}
where $k$ represents the modality of the visible or infrared images and $\Phi_{\mathcal{F}}$ is the paprameter of the feature extractor $\mathcal{F}(.)$. $C, H$ and $W$ means $C$ features with height $H$ and width $W$. In this manner, our feature extract network can significantly preserve spatial structure information for the common space and keep the cross-view consistency.

During the training process, each mini-batch data contains different modalities images simultaneously. Then our designed sharable dual-path information-preserving network extract 2048-dim feature from mini-batch data for further feature embedding. During testing, we use 2048-dim feature extracted from the pooling layer as final Re-ID features for different modalities images.

\subsection{Dual-Linear with Batch Normalization ID Embedding.} Identity embedding network (IDE) is a bastic a baseline in person Re-ID task. The last layer of IDE network, which outputs the ID prediction logits of images, is a fully-connected layer whose size of last hidden units is equal to numbers of person identities. Learning such a sharable single-linear classifier is widely used in visible-infrared person Re-ID problem, but this structure bring about a severe matter that the shared network is hard to converge with cross-modality ID embedding due to vanishing gradient, resulting in less discriminative cross-modality feature presentations. The main reason for this is that the feature maps from the global pooling are directly fed into the classifer with single fully-connected layer, causing the loss of detail and internal covariate shift.

To address above issue, we augment the classifier network by proposing the Dual-Linear with Batch Normalization ID Embedding (DL-IDE) method to further reduce cross-modality discrepancy for VI-ReID task. The illustration is shown in Fig. 5. Different from other sharable classifier, we add a new linear (fully-connected) layer after the pooling layer and a batch normalization layer before the last fully-connected layer of cross-modality ID embedding. The reason for positioning two linear layers, rather than the single one is that the model fails to learn identity discriminative classifier for two different modalities, which might be due to that the correlation of weight vectors in the last FC layer is determined by the training sample distribution. When eigenvectors of different modalities with the same identity are used for shared ID embedding, single-linear classifier has to get the average value of multiple patterns to fit, resulting in less discriminative cross-modality feature representations in the backward propagation learning process. Therefore, in order to make the modality dustributions more distinguishable, we utilize a new FC layer to model the subspace projection for modality alignment. 

\begin{figure}[t]
	\begin{center}
		\includegraphics[width=0.99\linewidth]{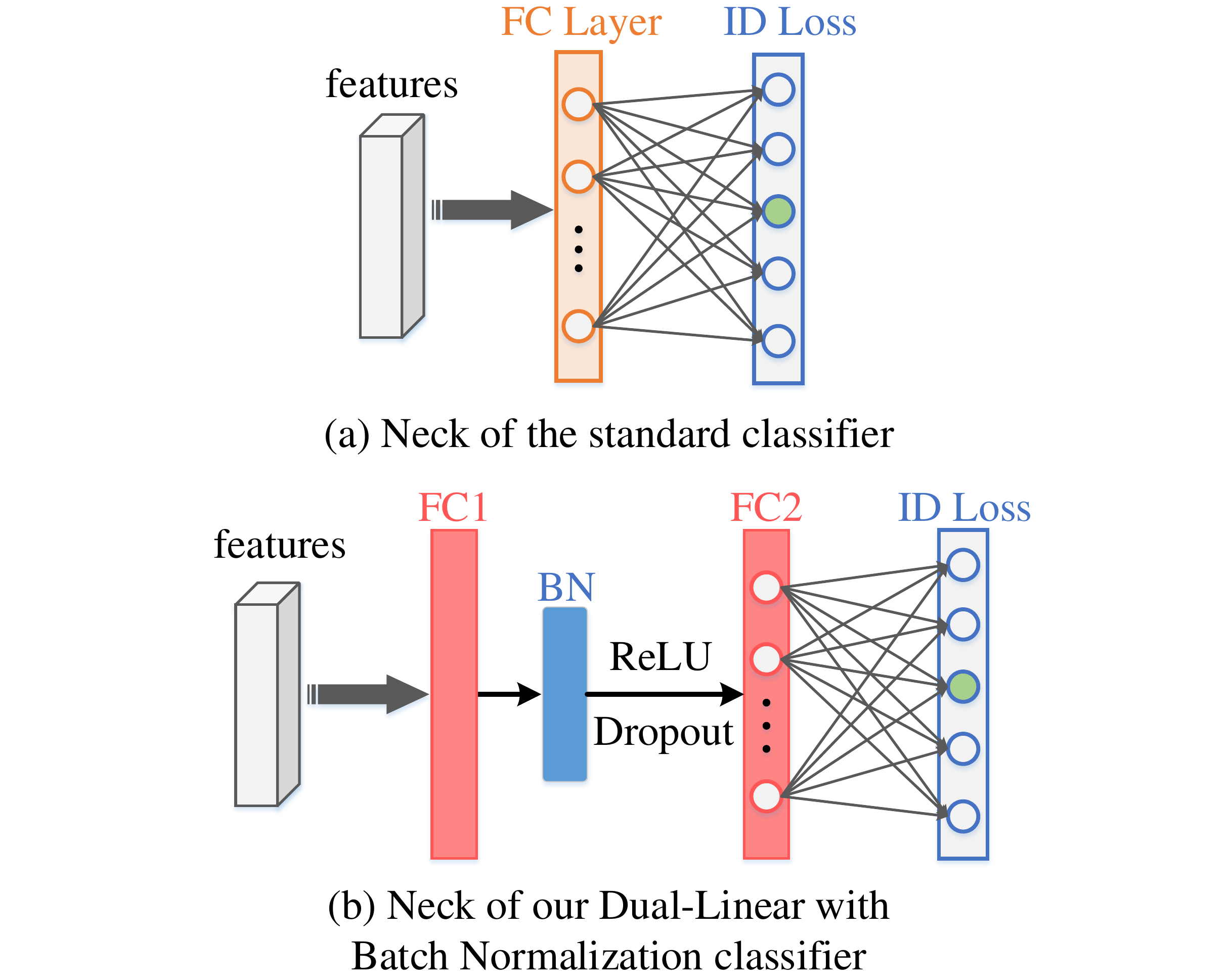}\vspace{-0.4cm}
	\end{center}
	\caption{Comparison between the widely-used classifier and our proposed dual-linear batch normalized classifier (DL-IDE). We used the outputs from the last fully-connected layer as the final classification results.\vspace{-0.3cm}}
	\label{fig:smalltarget}
\end{figure}

To clarify, we explain the limitation of the single-linear classifier from a theoretical point of view. Generally, the last linear layer (without bias term) calculates dot product between feature vectors $\left\{\overrightarrow{x}_i, i \in [1, C]\right\}$ and weight vectors $\left\{\overrightarrow{w}_i, i \in [1, C]\right\}$ of different classes. For each feature vector $\overrightarrow{x}_i$, we can compute output value by formula: $\overrightarrow{x}_i\overrightarrow{w}_j = |\overrightarrow{x}_i||\overrightarrow{w}_j|cos(\theta(i, j))$. Note that a good classification result mainly depends on the angles between two vectors $\left\{\theta(i, j), j \in [1, C]\right\}$ and the modulus of weight vectors $\left\{|\overrightarrow{w}_j|, j \in [1, C]\right\}$. However, for the neck of the standard classifier, the modulus of weight vectors $\left\{|\overrightarrow{w}_i|, i \in [1, C]\right\}$ should be similar to each other, leading to unsatisfied performance towards some particular classes from different modalities. By contrast, we add a new linear layer (with bias term) to project the feature vectors from different modalities into a more discriminative common feature space. Then, we utilize the last linear layer (without bias term) to produce classification result. A fully-connected layer with strong prior bias helps the identity loss in this way.

\textbf{Common Space Batch Normalization.} In most of conventional Re-ID models, the feature vectors from the pooling layer are directly used for identity classification, resulting in poor generalization and slow convergence. Inspired by \cite{30}, we propose to apply a weighting-sharing 1-D BN layer on the common feature embedding space to balance different losses in magnitude stabilizing. For the feature embedding vector $f_{initial}$, the normalized vector $\widehat{f}$ is calculated as:
\begin{equation} 
\widehat{f}=\frac{f_{initial} - E[f_{initial}]}{\sqrt{Var[f_{initial}]}},
\end{equation}
where $E[.]$ means the value of expectation and $Var[.]$ denotes the value of statistical variance.

The trainable parameters $\phi = \sqrt{Var[f_{initial}]}$ and $\varphi = E[f_{initial}]$ are the scaling and shift paprameters to ensure the batch normalization can degrade to the identity transformation. Note that $\varphi$ is dislodged from the common space batch normalization operator, the final embedding vector can be obtained by:
\begin{equation} 
f^{final} = \phi\widehat{f}.
\end{equation}

With the extra scaling and shift paprameters in batch normalization layer, the input of the last fully-connected layer are transformed into a form that ensures zero mean and unit variance, which helps recalibrate the channels of embedding vectors.

\textbf{Cross-Modality Sharable Identity Loss.} Define set of training images by $X^g$ and $X^t$ with identity labels $Y = \left\{y_i\right\}$, where $X^g$ represent grayscale training samples ($X^g = \left\{x_i^g|i = 1, ,2, ..., N_1\right\}$) and $X^t$ denote themal (infrared) training samples ($X^t = \left\{x_i^t|i = 1, ,2, ..., N_2\right\}$). Note that $N_1$ and $N_2$ mean the number of grayscale and infrared images in the training set respectively. In our learning process, we random select $n$ grayscale and $n$ infrared images to construct the batch. Given a grayscale sample $x_i^g$ with identity label $y_j$, we use a softmax function to calculate the its probability $p({y_{i}|x_i^v})$, formulated as:
\begin{equation} 
p({y_{i}|x_i^g})=\frac{exp(W^T_{g,j}f_g)}{\sum_{n=1}^{C}exp(W^T_{g,n}f_g)}, n = 1, 2, ..., C,
\end{equation}
where $f_g$ represent the feature fed into the dual-linear with batch normalization ID embedding module. $W_{g,j}$ is the weight matrix of the last fully connected layer in our classifier for $j$th identity. $C$ is the total number of identities.

Similarly, for a infrared sample $x_i^t$ with identity label $y_j$, we have:
\begin{equation} 
p({y_{i}|x_i^t})=\frac{exp(W^T_{t,j}f_t)}{\sum_{n=1}^{C}exp(W^T_{t,n}f_t)}, n = 1, 2, ..., C.
\end{equation}

With the calculated probabilities, the cross-entropy loss is used to optimize the dual-linear with batch normalization ID embedding, denoted by:
\begin{equation} 
\mathcal{L}_{id}=-\frac{1}{n}\sum_{f_k\in\left\{G, I\right\}}log(p({y_{i}|x_{i}^k})),
\end{equation}
where $G, I$ denote the features are extracted from grayscale and infrared images and $n$ represents the number of grayscale (infrared) samples at each training batch.

\subsection{Learning with Bi-directional Tri-constrained Top-Push Ranking Loss.} We propose a novel bi-directional tri-constrained top-push ranking loss to guide the feature learning process. It mainly contains threefold optimized objectives, namely cross-modality, intra-modality and inter-modality regularization constraints. Our solution is the improved version of \cite{12} which introduced the bi-directional dual-constrained top-ranking loss that includes cross-modality and intra-modality regularizers. In what follows we first revisit the bi-directional ranking loss, and then deduce our proposed bi-directional tri-constrained top-push ranking loss.

\subsubsection{Revisit Bi-directional Ranking Loss.} The bi-directional ranking loss is designed for cross-modality Re-ID task that considers two kinds of relationships: visible to infrared (sampling the anchor term from RGB domain, sampling the positive and negative terms from infrared domain) and infrared to visible (sampling the anchor term from infrared domain, sampling the positive and negative terms from RGB domain). Given a mini-batch that contains $N$ visible images and $N$ infrared images, the training samples from two different modalities can be defined as $X^v=\left\{x_i^v \right\}_{i=1}^N$  and $X^t=\left\{x_i^t \right\}_{i=1}^N$ with corresponding labels $Y^v=\left\{y_i^v \right\}_{i=1}^N$ and $Y^t=\left\{y_i^t \right\}_{i=1}^N$. The bi-directional ranking loss includes visible anchor based term $\mathcal{L}_{triplet}^{V}$ and infrared anchor based term $\mathcal{L}_{triplet}^{I}$ can be denoted by:
\begin{equation}
\begin{split}
\mathcal{L}_{bi\_rank} & =  \mathcal{L}_{triplet}^{V} + \mathcal{L}_{triplet}^{I} \\
=& \frac{1}{N}\sum_{\forall{y_i=y_j,y_i\neq y_k}}max[\zeta + D(x_i^v,x_j^t) - D(x_i^v,x_k^t), 0] \\
+& \frac{1}{N}\sum_{\forall{y_i=y_j,y_i\neq y_k}}max[\zeta + D(x_i^t,x_j^v) - D(x_i^t,x_k^v), 0],
\end{split}
\end{equation}
where $\zeta$ is a margin parameter. The subscripts $i$ and $j$ means the same label, while $i$ and $k$ are different labels.

\subsubsection{Bi-directional Tri-constrained Top-Push Ranking Loss.} To adapt the feature discrepancies of cross-modality and intra-modality, we improve the metric learning method with proposing a strong bi-directional tri-constrained top-push ranking loss for cross-modality Re-ID task. We employ three-fold triplet loss to optimize the network, i.e., cross-modality top-push ranking loss, intra-modality top-push ranking loss and inter-modality top-push ranking loss. Fig. 6 exhibits the relationship among heterogeneous images (grayscale modality and infrared modality in this work).

\textbf{Cross-Modality Top-Push Ranking Loss.} Considering the grayscale-infrared cross-modality matching protocol during testing, a cross-modality top-push ranking loss is first designed to enhance the feature discriminability with positive grayscale-infrared pairs and negative infrared-grayscale pairs. We use the formula $D(x_{i}^{g},x_{j}^{t})$ to represent the Euclidean distance between two samples $x_{i}^{g}$ and $x_{j}^{t}$, where the superscript $g$ means grayscale modlaity, $t$ means themal (infrared) modality and the subscripts $\left\{ i, j \right\}$ represent the image index. The cross-modality top-push ranking loss can be written as: 
\begin{equation} 
\begin{split}
\mathcal{L}_{cross}=&\sum_{i=1}^{N}[\delta_1 +  \mathop{max}\limits_{\forall{y_{i}=y_{j}}}D(x_{i}^{g},x_{j}^{t+}) - \mathop{min}\limits_{\forall{y_{i}\neq y_{k}}}D(x_{i}^{g},x_{k}^{t-})]_{+}  \\
+& \sum_{i=1}^{N}[\delta_1 +  \mathop{max}\limits_{\forall{y_{i}=y_{j}}}D(x_{i}^{t},x_{j}^{g+}) - \mathop{min}\limits_{\forall{y_{i}\neq y_{k}}}D(x_{i}^{t},x_{k}^{g-})]_{+},
\end{split}
\end{equation}
where $[ . ]_{+}$ represents $max(0,x)$ function to guarantee the non-negativity constraint, $\delta$ is a pre-defined margin , and $N$ means the number of infrared or grayscale inputs in each training batch.

\textbf{Intra-Modality Top-Push Ranking Loss.} VI-ReID also suffers from intra-class variations caused by distinct viewpoints, pose variations and scale changes. To address this problem, another top-push ranking loss is employed for each modality to reduce the feature distances between images of the same person and enlarge the distances between images of different people. Based on the cross-modality constrained loss, the intra-modality top-push ranking loss for grayscale and infrared modalities is represented by:

\begin{figure}[t]
	\begin{center}
		\includegraphics[width=0.99\linewidth]{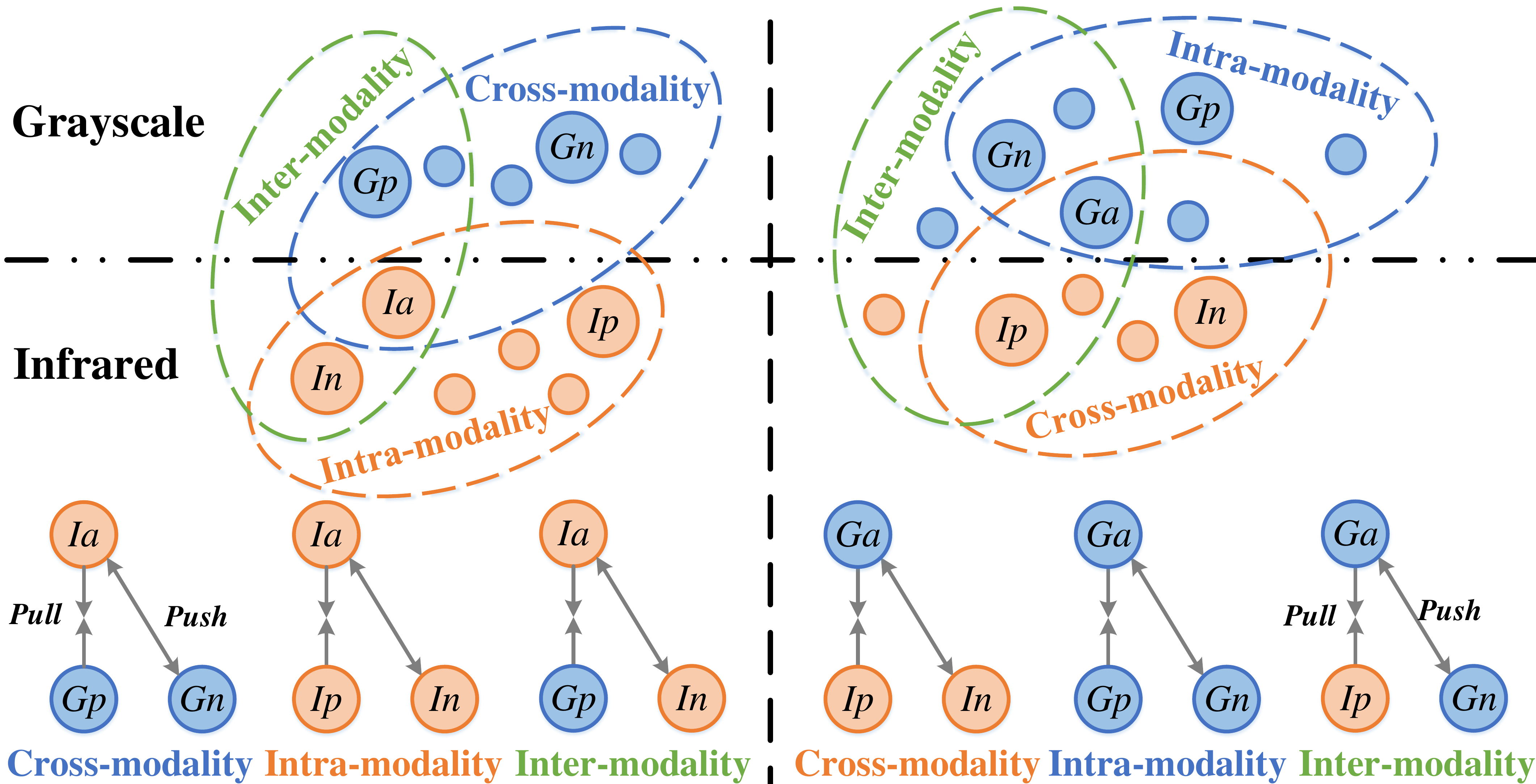}\vspace{-0.4cm}
	\end{center}
	\caption{Illustration of the distance relationship between grayscale images $G$ and infrared images $I$. $a$, $p$, $n$ represent the anchor, positive and negative samples, repectively. We use three regularization constraints to pull close those samples with the same label and push far away those samples with different labels regardless of which modality is from.\vspace{-0.3cm}}
	\label{fig:smalltarget}
\end{figure}

\begin{equation} 
\begin{split}
\mathcal{L}_{intra}=&\sum_{i=1}^{N}\mathop{max}\limits_{\forall{y_{i}=y_{j}}}[\delta_2 +  D(x_{i}^{g},x_{j}^{g+}) - \mathop{min}\limits_{\forall{y_{i}\neq y_{k}}}D(x_{i}^{g},x_{k}^{g-}),0] \\+ &\sum_{i=1}^{N}
\mathop{max}\limits_{\forall{y_{i}=y_{j}}}[\delta_2 +  D(x_{i}^{t},x_{j}^{t+}) - \mathop{min}\limits_{\forall{y_{i}\neq y_{k}}}D(x_{i}^{t},x_{k}^{t-}),0].
\end{split}
\end{equation}
The intra-modality top-push ranking loss ensures that the closest intra-modality negative sample should be far from the farthest intra-modality positive sample.

\textbf{Inter-Modality Top-Push Ranking Loss.} To further minimize the cross-modality varitions, we add a novel optimizied regularizer to bi-directional tri-constrained top-push ranking loss. It significantly considers the distance relationship between cross-modality and intra-modaltiy and ensures that the learnt feature is modality-invariant for metric learning, which is formulated by:
\begin{equation} 
\begin{split}
\mathcal{L}_{inter}=&\sum_{i=1}^{N}\mathop{max}\limits_{\forall{y_{i}=y_{j}}}[\delta_3 +  D(x_{i}^{g},x_{j}^{t+}) - \mathop{min}\limits_{\forall{y_{i}\neq y_{k}}}D(x_{i}^{g},x_{k}^{g-}),0] \\+ &\sum_{i=1}^{N}
\mathop{max}\limits_{\forall{y_{i}=y_{j}}}[\delta_3 +  D(x_{i}^{t},x_{j}^{g+}) - \mathop{min}\limits_{\forall{y_{i}\neq y_{k}}}D(x_{i}^{t},x_{k}^{t-}),0].
\end{split}
\end{equation}
For a grayscale anchor sample $x_{i}^{g}$, note that we choose the farthest grayscale-infrared positive pair and the closest grayscale-grayscale negative pair to formulate a mined informative triplet $\left\{x_i^g, x_i^{t+}, x_k^{g-} \right\}$. This manner fully utilizes the cross-modality triplet-wise relationship and improve the robustness against modality varitions.

\textbf{Discussion of Dual-Constrained Top-Ranking and Tri-Constrained Top-push Ranking.} Triplet constraint is usually utilised in the classical ranking scheme. Dual-constrained top-ranking constraints contain a cross-modality triplet regularizer and a intra-modality triplet regularizer, which connect embedding vectors from different domains and resolve the discrepancy between different modalities. However, in this situation, inter-class feature differences are still ambiguous in the cross-modality setting. 

In comparison, the proposed tri-constrained top-push ranking loss adds a inter-modality regularizer to force intra-class difference to be smaller than the inter-class difference regardless of which modality is from. This modification has two major advantages: (a) further mitigate the data biases across modalities in the common space, (b) keep the powerful feature representation ability as well as the discriminative ability of top-push ranking constraint.

\textbf{Overall Embedding Loss.} Finally, the optimization $\mathcal{L}_{total}$ of the proposed grayscale-infrared feature learning framework is defined by the combination of cross-modality sharable identity loss ($\mathcal{L}_{id}$) and bi-directional tri-constrained top-push ranking loss ($\mathcal{L}_{cross}$, $\mathcal{L}_{intra}$, $\mathcal{L}_{inter}$):

\begin{equation} 
\mathcal{L}_{total}=\mathcal{L}_{id} + \mathcal{L}_{cross} + \lambda_1\mathcal{L}_{intra} + \lambda_2\mathcal{L}_{inter},
\end{equation}
where $\lambda_1$ and $\lambda_2$ are the predefined tradeoff parameters to balance the different losses.

The identity loss ($\mathcal{L}_{id}$) optimizes network parameters with the identity supervision, which encourages indentity-invariant feature representations. The bi-directional tri-constrained top-push ranking loss (BTTR) provides a strong supervision to optimize the bi-directional relationship among different person images across the two modalities. 

\section{Experiments}

In this section, we evaluate the effectiveness of our proposed approach, including description of the experimental settings, ablation study of each component and comparsion with state-of-the-art methods. Experimental results are reported to answer two questions: 1) how do the components contribute to the performance? 2) What is the effect of our proposed framework when comparing with peer state-of-the-art methods?

\subsection{Experimental Settings}
\textbf{Datasets.} We mainly conduct our proposed method on the RegDB dataset \cite{27} and SYSU-MM01 \cite{5} dataset, which is a large-scale VI-ReID dataset in existing literatures.

SYSU-MM01 \cite{5}. This dataset was captured by six disjoint cameras (four general RGB and two near-infrared) in both indoor and outdoor environments and contains a total of 491 identities with 287,628 visible images and 15,792 infrared images. The training set contains images of 395 persons with 22,258 visible images and 11,909 infrared images, and the testing set contains 96 persons. Following Wu et al.\cite{5}, we also employ two test modes in our evaluation protocol, the all-search mode with all images and the indoor-search mode with only indoor images from cameras 1, 2, 3, and 6. 

RegDB \cite{27}. This dataset was collected by dual camera systems (with optical and thermal sensors) and contains a total of 412 identities. Each identity has 10 different visible images and infrared images. We follow the evaluation protocol in \cite{27} to randomly divide the dataset into two halves, one half for training and the other half for testing. The training subset consists of 2,060
visible images and 2,060 infrared images. This also applies to the testing set. We evaluate the performance via changing the query setting to visible (query) to thermal (gallery).

\textbf{Evaluation Protocol.} In the evaluation stage, we utilize the Cumulated Matching Characteristics (CMC) curve and mean Average Precision (mAP) evaluation criteria to evaluate our proposed method. CMC is adopted to report the results of rank-n accuracy, which denotes the probabilities that a query object appears in the target lists. The metric mAP computes the average value of the maximum recalls for each class in multiple types of tests, which can reflect the overall ranking accuracy \cite{28}. Note that there is a slight difference from the single-modality Re-ID problem. In VI-ReID issue, images of one modality act as the query set while the images from the other modality act as the gallery.

\begin{table}[t]
	\centering
	\caption{Comparison with GAN-based methods using the same backbone and regularization losses (ID+Triplet) on the SYSU-MM01 dataset, including both all-search and indoor-search modes.}
	\label{Tab03}
	\begin{tabular}{l|cc|cc}	
		\hline
		& \multicolumn{2}{c}{All Search} & \multicolumn{2}{c}{Indoor Search} \\	
		\hline			
		{Mehod}	&Rank-1  &mAP  &Rank-1  &mAP\\			
		\hline	
		RGB-GAN (CoSiGAN)       & 35.55& 38.33 & -  &-  \\	
		GAN-Infrared (AlignGAN)    & 42.40&40.70&-&-  \\
		GAN-GAN \cite{16}       & 46.80&38.00&46.80&54.70 \\
		Grayscale-Infrared (Ours)  & \textbf{55.17}&  \textbf{51.38} & \textbf{59.56}  & \textbf{67.29}  \\
		\hline		
	\end{tabular}
\end{table}

\begin{table*}
	
	\caption{Evaluation of each component on the large-scale SYSU-MM01 dataset. `grayscale' means utilizing grayscale modality augementation strategy. We strart using the RGB-Infrared learning strategy (Baseline) and Grayscale-Infrared learning strategy (Grayscale), and then gradually add other components for evaluation.}
	\label{tab:1}       
	\begin{center}
		\begin{tabular}{l|ccccc|ccccc}
			\hline
			Modes & \multicolumn{5}{c}{\emph{All Search}} & \multicolumn{5}{c}{\emph{Indoor Search}} \\
			\hline
			Method &Rank-1 &Rank-5 &Rank-10 &Rank-20 &mAP &Rank-1 &Rank-5 &Rank-10 &Rank-20 &mAP \\
			\hline
			SFANet (RGB) with Baseline &38.81 &70.02 &80.65 &89.32 &35.30 &44.21 &76.19 &87.13 &93.93 &50.20  \\
			SFANet (RGB) + DL-IDE &48.96 &75.89 &83.99 &92.96 &48.23 &53.59 &81.08 &90.18 &96.60 &59.78  \\
			SFANet (RGB) + BTTR &45.96 &72.89 &81.99 &89.96 &46.23  &51.59 &79.08 &89.18 &96.60 &57.78 \\
			SFANet (RGB) + DL-IDE + BTTR &52.56 &77.54 &85.77 &92.11 &50.93  &54.17 &82.17 &90.17 &94.66 &61.84 \\
			\hline
			\hline
			SFANet (Gray) with Baseline  &43.70 &69.50 &79.15 &87.06 &40.36 &50.05 &81.07 &90.81 &96.24 &56.66 \\
			SFANet (Gray) + DL-IDE  &60.45 &84.01 &91.80 &95.16 &53.87 &66.57 &89.15&93.47&98.35&75.95 \\
			SFANet (Gray) + BTTR &58.35 &82.83 &90.24 &94.85 &53.00 &64.80 &88.69 &94.67 &98.07 &75.16 \\
			SFANet (Gray) + DL-IDE + BTTR  &65.74 &87.93 &92.98 &97.05 &60.83 &71.60 &91.08 &96.60 &99.45 &80.05 \\
			\hline
		\end{tabular}
	\end{center}	
\end{table*}

\textbf{Implementation Details.} We implement our proposed method on Pytorch framework and use NVIDIA TITAN RTX graphics card for acceleration. Following most existing person Re-ID works, Resnet-50 \cite{29} is used as backbone for feature extraction. All input images are first resized to 288 $\times$ 144. We initialize the convolutional blocks with weights pre-trained on ImageNet. The dimensions of the last classification layer are 395 for SYSU-MM01 and 206 for RegDB, respectively. The training samples are augmented with two methods, random horizontal flipping and random cropping. The total number of training epochs is 80, and the batch size is setted to 8. We start training with learning rate 0.01 and linearly increase to 0.1 in the first 10 epochs, then, we keep the same value setting until researching to 20 epochs. In the following 60 epochs, learning rate is set to 0.01 for the first 30 epochs and 0.001 for another 30 epochs. We adopt the SGD optimizer with a weight decay of 5$\times$$10^{-4}$ and a momentum of $0.9$ to update the parameters of the network. We set the margin parameter $\delta_1$ to 0.5 in Eq. 10, $\delta_2$ in Eq. 9 to 0.1 and $\delta_3$ in Eq. 11 to 0.3. The tradeoff parameter $\lambda_1$ is set to 0.1 and $\lambda_2$ is set to 0.5. In testing, feature distance is calculated by Euclidean metric.

\subsection{Ablation Study}
In this subsection, we evaluate the effectiveness of each component of our proposed SFANet. We select the all-search and indoor-search modes of SYSU-MM01 \cite{5} for ablation study. Specifically, `SFANet (RGB)  with Baseline' means the baseline results by using the initial visible and infrared images as input tensors to be fed into the network, which is trained with standard identity loss $\mathcal{L}_{id}$ and triplet loss $\mathcal{L}_{tri}$. `SFANet (Gray) with Baseline' denotes the results obtained with the converted grayscale and 3-channel infrared images as network inputs and other settings are same as the `RGB'. `DL-IDE' represents the results obtained by using dual-linear with batch normalization ID embedding method. `BTTR' demonstrates the bi-directional tri-constrained top-push
ranking loss with hard mining.

\begin{table}[t]
	\centering
	\caption{Ablation study of the proposed dual-linear with batch normalization ID embedding method.}
	\label{Tab03}
	\begin{tabular}{l|cc|cc}	
		\hline
		& \multicolumn{2}{c}{All Search} & \multicolumn{2}{c}{Indoor Search} \\	
		\hline			
		{Setting}	&Rank-1  &mAP  &Rank-1  &mAP\\			
		\hline		
		Baseline    &43.70&40.36&50.05&56.66  \\
		Baseline + BN      &54.96&51.95&59.74& 64.91  \\
		Baseline + Dual-Linear  &46.14&42.08 &54.23  & 59.01  \\
		Baseline + Dual-Linear + BN  &60.45&53.87 &66.57  &75.95  \\
		\hline		
	\end{tabular}
\end{table}

\begin{figure}[t]
	\begin{center}
		\includegraphics[width=0.99\linewidth]{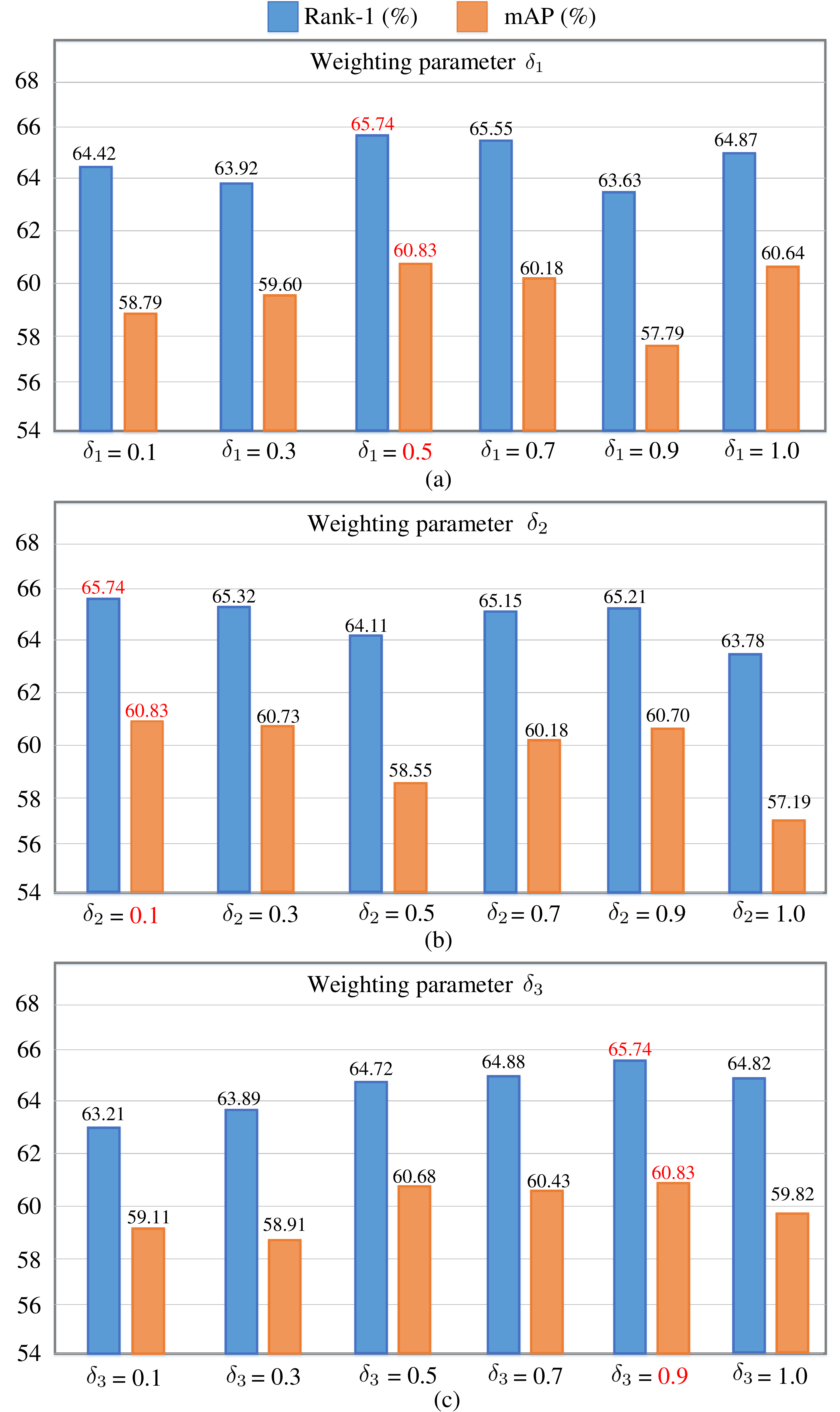}\vspace{-0.4cm}
	\end{center}
	\caption{Performance of BTTR with different values of margins, \textit{i.e.}, $\delta_1$, $\delta_2$ and $\delta_3$ in cross-modality, intra-modality and inter-modality regularization terms repectively. We set $\delta_1$ = 0.5 and $\delta_2$ = 0.1 for evaluating margin parameter $\delta_3$ $\in$ (0,1] in the bottom plot. Similarly, we also set $\delta_2$ = 0.1 and $\delta_3$ = 0.9 for evaluating margin parameter $\delta_1$ $\in$ (0,1] in the top plot. And the same evaluation protocol for $\delta_2$. All the evaluations are conducted on the SUSY-MM01 dataset under all-search mode.  \vspace{-0.3cm}}
	\label{fig:smalltarget}
\end{figure}

\textbf{Effectiveness of Grayscale-Spetrum Images.} Our solution is based on grayscale-infrared feature learning method. To show the effect of this strategy, We conduct experiments with other three different GAN-based solutions in Table \Rmnum{1}. Specifically, `RGB-GAN' means exploiting GAN to generate RGB images from the infrared modality. `GAN-Infrared' denotes generating infrared images from RGB modality. `GAN-GAN' represents the cross-modality paired-images generation. For experimental fairness, our grayscale-infrared learning method is conducted on a one-stream network under the supervision of a standard identity loss and a triplet loss, which is similiar to the other three methods. From the Table \Rmnum{2}, we can observe that our graysacle-spetrum feature augementation strategy achieve a siginificant performance improvement over GAN-based methods on both all-search and indoor-search modes.

In addition, to fully present the effectiveness of VI-ReID solution method based on spetrum-aware feature augmentation, we also conduct an ablation experiment to see how the performance will change if we use conventional RGB-infrared learning strategy. The results are reported in Table \Rmnum{2}. As we can see in Table \Rmnum{2}, without grayscale-spetrum image augmentation, the result in the first row is inferior where rank-1 accuracy is 38.81$\%$ and mAP score is 35.30$\%$. In comparison, our grayscale augmentation method presented in the fifth row achieves 43.70$\%$ rank-1 accuracy and 40.36$\%$ mAP score, which improves the performance by a large margin. 

\begin{table}[t]
	\centering
	\caption{Effectiveness of different regularization terms of BTTR on the SYSU-MM01 dataset. Re-identification rates ($\%$) at rank-1 and mAP ($\%$) are reported.}
	\label{Tab03}
	
	\begin{tabular}{l|cc|cc}	
		\hline
		& \multicolumn{2}{c}{All Search} & \multicolumn{2}{c}{Indoor Search} \\	
		\hline			
		{Methods}	&Rank-1  &mAP  &Rank-1  &mAP\\			
		\hline		
		$\mathcal{L}_{id}$ \textit{with DL-IDE}   &60.45&53.87&66.57&75.95  \\
		$\mathcal{L}_{id}+\mathcal{L}_{cross}$      &62.61&56.11 &68.32  &78.41  \\
		$\mathcal{L}_{id}+\mathcal{L}_{intra}$     & 61.03&54.17 &67.32  &76.51  \\
		$\mathcal{L}_{id}+\mathcal{L}_{inter}$     & 61.75&55.22 &67.99  &77.88  \\
		$\mathcal{L}_{id}+\mathcal{L}_{cross}+\mathcal{L}_{intra}$ \textit{(BDTR)}     & 63.33&58.22 &69.17  &78.57  \\
		$\mathcal{L}_{id}+\mathcal{L}_{cross}+\mathcal{L}_{intra}+\mathcal{L}_{inter}$  & \textbf{65.74}&  \textbf{60.83} & \textbf{71.60}  & \textbf{80.05}  \\
		
		\hline		
	\end{tabular}
	
\end{table}

\textbf{Effectiveness of Dual-Linear with Batch Normalization ID Embedding Method.} To better understand the property of dual-linear with batch normalization ID embedding method, we study the effectiveness of each component of it. Here, `Baseline' means that only one fully-connected layer are used for classification and do not use triplet loss for optimization, `BN' represents batch normalization operation, and `Dual-Linear' means the classifier contains two fully-connected layer. As shown in Table \Rmnum{3}, when we use batch normalization operation on baseline, the network achieves competitive performance with improving the baseline by 9$\%$$\sim$11$\%$. This is because `BN' reshapes the discribution of embedding vectors and smoothen the optimization landscape. Then we add a new fully-connected layer which projects the Re-ID feature vectors into a more discriminative classification space, the performance of rank-1 and mAP increase by 3$\%$$\sim$4$\%$. It is worth to point that if we combine `BN' and `Dual-Linear' strategy, the performance can be further improved to 60.45$\%$(rank-1), 53.87$\%$(mAP).

\textbf{Effectiveness of Bi-directional Tri-constrained Top-Push Ranking Loss.}
We also design a series of experiments based on our proposed dual-path network to test the effectiveness of BTTR. Since our bi-directional tri-constrained top-push ranking loss contains three parts, we test the influence of each component respectively. Note that we compare performance with optimizied regularization terms to the cases in which each regularizer is used individually. The result is reported in Table \Rmnum{4}. From the Table, we can observed that $\mathcal{L}_{cross}$ improves the performance of model by 2$\%$$\sim$3$\%$, which verfies that cross-modality constraint can effectively handles the modality varitions between two different modalities. With $\mathcal{L}_{intra}$ or $\mathcal{L}_{inter}$, the performance have a slightly improved by 1$\%$$\sim$2$\%$. This observation demonstrates that intra-modality and inter-modality top-push ranking losses provide the relatively weaker constraints for cross-modality metric, but still can improve feature learning consistently. 

Another observation is that the proposed BTTR achieves a siginificant improvement than BDTR since the new introduced inter-modality regularizer considers the distance relationship between cross-modality and intra-modaltiy. Furthermore, it is worth to point that when we combine three optimizied regularization terms (BTTR), rank-1 accuracy and mAP increase from 60.45$\%$ and 53.87$\%$ to 65.74$\%$ (\textbf{+5.29$\%$}) and 60.83$\%$ (\textbf{+6.96$\%$}), which improve the performance by a large margin.  

\textbf{Hyperparametric analysis.} \textit{1) Weighting Parameters:} We analyze the contributions of the weighting parameters $\lambda_1$ and $\lambda_2$ on SYSU-MM01 dataset. Specifically, $\lambda_1$ controls the contribution of the intra-modality regularization and $\lambda_2$ controls the contribution of the inter-modality regularization. We vary their value from 0 to 1 and evaluate the resulting performance on SYSU-MM01. From the Fig. 8, we can observe that the Re-ID performance rise with increasing of $\lambda_1$ and $\lambda_2$ and the best performance is acheived with $\lambda_1$ = 0.1, $\lambda_2$ = 0.5. Empirically, to balance each part in the objective, we set parameters $\lambda_1$ = 0.1, $\lambda_2$ = 0.5 when using BTTR loss.
\begin{figure}[t]
	\begin{center}
		\includegraphics[width=0.99\linewidth]{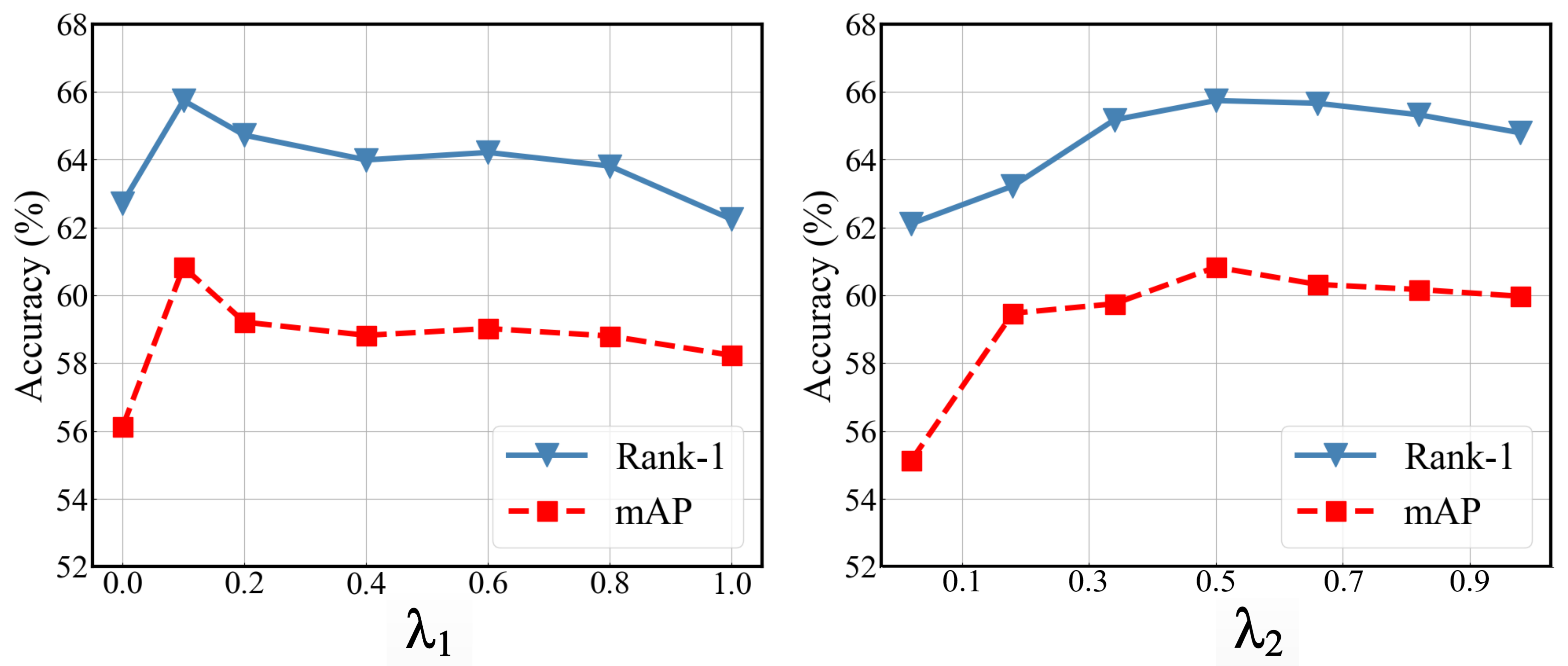}\vspace{-0.4cm}
	\end{center}
	\caption{Evaluation of the weighting parameter $\lambda_1$ and $\lambda_2$ on SYSU-MM01 dataset (left: $\lambda_2$ = 0.5 and $\lambda_1$ $\in$ [0,1], right: $\lambda_1$ = 0.1 and $\lambda_2$ $\in$ [0,1]).\vspace{-0.3cm}}
	\label{fig:smalltarget}
\end{figure}

\textit{2) Margin Parameters:} We also evaluate the margin parameters $\delta_1$, $\delta_2$ and $\delta_3$  of the $\mathcal{L}_{cross}$, $\mathcal{L}_{intra}$ and $\mathcal{L}_{inter}$ respectively. Fig. 8 shows the peformance changes under two evaluation metrics (rank-1 and mAP) on SYSU-MM01 dataset. According to the literature \cite{12}, we empirically assign a small value of 0.1 for $\delta_2$, the $\delta_1$ is set to 0.5 in our method. Actually, the results shown in Fig.7(a-b) also verifies the rationality of this setting. In this situation, from the Fig.7(c), we can observe that the performance fluctuate dynamicly with the different value of $\delta_3$ and model achieves peak performance when $\delta_3$ is equal to 0.9. This is because $\delta_3$ considers the distance relationship between cross-modality and intra-modaltiy. When we select a positive sample from the different modality and a negative sample from the same modality, the distance between positive and negative sample pairs usually becomes very close. Therefore, a relative large value for $\delta_3$ is beneficial for fast convergence of the network. In addition, this experiment also demonstrates that a suitable margin parameter is significant for cross-modality person re-identification task.

\begin{table*}[t]
	\caption{Comparison with the state-of-the-art methods under all-search and indoor-search modes on SYSU-MM01 dataset.}
	\label{tab:1}
	\begin{center}

		\begin{tabular}{l|c|cccc||ccccc}	
			\hline
			&	 & \multicolumn{4}{c}{\emph{All-search}} & \multicolumn{4}{c}{\emph{Indoor-Search}} \\
			
			\hline
			
			Method & Venue	& Rank-1      &  Rank-10   &   Rank-20   &  mAP
			
			& Rank-1      &  Rank-10   &   Rank-20   &  mAP
			\\
			\hline
			
			Two-Stream \cite{5}&ICCV2017  & 11.65&  47.99 &  65.50  & 12.85 & 15.60&  61.18  & 81.02  & 21.49 \\
			One-Stream \cite{5}&ICCV2017 & 12.04&  49.68  & 66.74  & 13.67 &16.94&63.55&82.10&22.95 \\
			Zero-Pad \cite{5}  &ICCV2017 &14.80&54.12&71.33&15.95 & 20.58& 68.38  & 85.79  & 26.92\\
			HCML \cite{11} &AAAI2018 & 14.32& 53.16  & 69.17  & 16.16 & 24.52& 73.25  & 86.73  & 30.08 \\
			cmGAN \cite{13} &IJCAI2018 & 26.97&  67.51  &  80.56  & 31.49 & 31.63& 77.23  & 89.18  & 42.19 \\
			eDBTR \cite{24} &TIFS2020 & 27.82&  67.34  &  81.34  & 28.42 & 32.46& 77.42  & 89.62  & 42.46 \\
			HSME \cite{35} &AAAI2019 & 20.68&  32.74  &  77.95  & 23.12 & -&  -  &  -  & -\\
			D$^{2}$RL \cite{14} &CVPR2019 & 28.90&  70.60  & 82.40  & 29.20 & -&  -  &  -  & - \\
			MAC \cite{36} &ACM MM2019 & 33.26&  79.04  & 90.09  & 36.22 & 36.43& 62.36 & 71.63  & 37.03 \\
			MSR \cite{23}  &TIP2020 & 37.35& 83.40 & 93.34  & 38.11 & 39.64&  89.29 & 97.66 & 50.88 \\
			AlignG \cite{10} &ICCV2019 & 42.40&  85.00  &  93.70  & 40.70 & 45.90&  87.60  & 94.40  & 54.30 \\
			CoSiGAN \cite{15} &ICMR2020 & 35.55& 81.54 & 90.43 & 38.33 &- &- & - &- \\
			X-Modal \cite{25} &AAAI2020 & 49.92 & 89.79 & 95.96 & 50.73 &- &- & - &- \\
			HPILN \cite{31} &ArXiv2019 & 41.36 & 84.78 & 94.31 &42.95 &45.77 &91.82 &98.46 &56.52 \\
			LZM \cite{32} &ArXiv2019 & 45.00 & 89.06 &- & 45.94 &49.66 &92.47 & - &59.81 \\
			MACE \cite{33} &TIP2020 & 51.64 & 87.25 & 94.44 &50.11 &57.35 &93.02 &97.47 &64.79 \\
			DDAG \cite{34} &ECCV2020 & 54.75 & 90.36 & 95.81 & 53.02 &61.02 &94.06 &98.41 &67.98\\
			cm-SSFT \cite{44} &CVPR2020 & 61.60 &89.20 & 93.90 &\textbf{63.20} &70.50 &94.90 &97.70 &72.60\\
			\hline
			\hline
			SFANet (Ours) &-- & \textbf{65.74}&\textbf{92.98}  & \textbf{97.05}  &60.83 &\textbf{71.60}&\textbf{ 96.60}   &\textbf{99.45}  & \textbf{80.05}\\
			
			\hline
			
		\end{tabular}
	\end{center}

\end{table*}
\begin{table}
	
	\caption{Comparison with the state-of-the-art methods on the RegDB datasets of different query settings.}
	\label{tab:1}       
	\begin{center}
		\begin{tabular}{l|cccc}
			\hline
			Setting & \multicolumn{4}{c}{\emph{Visible-Thermal}} \\
			\hline
			Method&Rank-1 &Rank-10 &Rank-20 &mAP \\
			\hline
			Zero-Pad \cite{5} &17.75 &34.21 &44.35 &18.90 \\
			HCML \cite{11}  &24.44 &47.53 &56.78 &20.80  \\
			BDTR \cite{12} &33.47 &58.42 &67.52 &31.83 \\
			eBDTR \cite{24}  &34.62 &58.96 &68.72 &33.46  \\
			HSME \cite{35}  &50.85 &73.36 &81.66 &47.00  \\
			D$^2$RL \cite{14}  &43.40 &66.10 &76.30 &44.10 \\
			MAC \cite{36}  &36.43 &62.36 &71.63 &37.03  \\
			CoSiGAN \cite{15}  &47.18 &65.97 &75.29 &46.16 \\
			MSR \cite{23}  &48.43 &70.32 &79.95 &48.67  \\
			EDFL \cite{45}  &48.43 &70.32 &79.95 &48.67  \\
			AlignG \cite{10}  &57.92 &- &- &53.64  \\
			DDAG \cite{34}  &69.34 &86.19 &91.49 &63.46  \\
			\hline
			SFANet (Ours) &\textbf{76.31} &\textbf{91.02} &\textbf{94.27} &\textbf{68.00} \\
			\hline
			\hline
			Setting & \multicolumn{4}{c}{\emph{Thermal-Visible}} \\
			\hline
			Zero-Pad \cite{5} &16.63 &34.68 &44.25 &17.82 \\
			HCML \cite{11}  &21.70 &45.02 &55.58 &22.24  \\
			eBDTR \cite{24}  &34.21 &58.74 &68.64 &32.49  \\
			HSME \cite{35}  &50.15 &72.40 &81.07 &46.16  \\
			MAC \cite{36}  &36.20 &61.68 &70.99 &36.63  \\
			EDFL \cite{45}  &51.89 &72.09 &81.04 &52.13  \\
			CDP \cite{46}  &64.40 &84.50 &90.80 &61.50  \\
			\hline
			SFANet (Ours) &\textbf{70.15} &\textbf{85.24} &\textbf{89.27} &\textbf{63.77} \\
			\hline
		\end{tabular}
	\end{center}
	
\end{table}

\subsection{Comparison with State-of-the-Arts}
We compare the performance of our proposed approach (GMA) with other state-of-the-art VI-ReID methods on SYSU-MM01 and RegDB in Table \Rmnum{5} and Table \Rmnum{7}, respectively. A variety of existing VI-ReID models are included for comparision, containing the up-to-date approachs of X-Modal \cite{25}, eDBTR \cite{24}, MSR \cite{23}, CosiGAN \cite{15}, MACE \cite{33}, DDAG \cite{34}, and cm-SSFT \cite{34} that published in 2020. In addition, some learning-based classic VI-ReID methods are also used for comparisons, such as Zero-Pad \cite{5}, HSME \cite{35}, cmGAN \cite{13}, D$^2$RL \cite{14}, HPILN \cite{31}, LZM \cite{32} and AlignG \cite{10}.




\textit{1) Comparisons on SYSU-MM01 Dataset:} We report the results of comparision with the state-of-the-art approaches using SYSU-MM01 dataset in Table \Rmnum{5}. Note that our proposed method achieves very satisfactory recognition accuracies in cross-modality matching task: the rank-1 acuuracy of 65.74$\%$, which is 10.99$\%$ absolute points higher than the recent DDAG \cite{34} algorithm. We also achieve an mAP of 60.83$\%$, outperforming most of the state-of-the-art methods but performing worse than cross-modality shared-specific feature transfer (cm-SSFT) method \cite{44}. This is because cm-SSFT utilizies both shared and specific information for each sample while our feature learning method with parameter sharing strategy chooses to abandon some modality specific cues to achieve a more complete spatial structure information of person in the feature extraction stage, causing a little performance fluctuation.  Compared to GAN-based methods like AlignG \cite{10} and D$^2$RL \cite{14}, our method achieves much higher accuracy on both CMC and mAP evaluation protocols and does not train an extra GAN model with high computional resources. In addition, compared to HCML \cite{11} which jointly optimizes the modality-specific and modality-shared metrics with human intervention, while our proposed method can be trained in an end-to-end way without interruption.

\textit{2) Comparisons on RegDB Dataset:} We also report the comparison results using RegDB dataset in Table \Rmnum{6}. Our method achieves outstanding recognition performance in both query settings. We outperform the best rank-1 accuracy by 6.25$\%$ and the best mAP by 2.27$\%$ (CDP) in thermal-to-visible setting. Moreover, we also improve the performance of the DDAG algorithm under the visible-to-thermal setting, from 69.34$\%$ to 76.31$\%$ in rank-1 accuracy and from 63.46$\%$ to 68.00$\%$ in mAP. This experiment demonstrates that our proposed method can alleviate the impact of severe modality discrepancy and learn discriminative feature representations via uilizing grayscale-spectrum augementation strategy.

\section{Discussion}
\subsection{Analysis of feature augmentation strategy via grayscale-spretrum images.} For VI-ReID problem, previous methods adopt RGB-Infrared feature learning strategy to eliminate the appearance and modality discrepancies. However, it is difficult to match the same person from different modalites in the unified space because visible and infrared images have an apparent cross-modality discrepancy and mapping such an RGB image to another heterogeneous modality may cause loss of channel information. 

Some GAN-based methods try to generate a modality-invariant representation for eliminating the modality discrepancy in pixel level. However, the generated images cannot possess good qualities to fill the modality gap between synthesized scenarios and target real ones. As shown in Fig. 9, we present the visualized results that utilize CycleGAN to synthesize fake infrared images. It can be observed that the synthetic images suffer from three main issues with image indistinctness, distorted body structure and addtional nosie. If we directly use these low-quality generated images to train Re-ID model, a novel gap between the original data and the synthetic data will be introduced to our learning process. In other words, although GAN-based method reduce the modality discrepancy in pixel-level, meanwhile it also enlarge the intra-person discrepancy that increases the difficulty of feature similarity matching. 

In comparison, the proposed method that expolits the grayscale-spectrum images to replace RGB images for feature learning is a great solution for this issue. It enjoys following serval metrits. First, the grayscale images can well preserve the key structural information of complete textures and semantic cues of visible images. They are critical for VI-ReID performance improvement since infrared images do not contain any visible information. Second, the grayscale modality can approximate the style of infrared images to the greatest extent. As a result, we can regard this subtle modality discrepancy between graycale and infrared images as a part of the appearance discrepancy that can be eliminated via using only feature-level constraints. From the Table \Rmnum{2}, it can be clearly seen that our grayscale-infrared learning method has a great performance improvement against RGB-Infrared training strategy. This experiment also provides a good suggestion for future VI-ReID research.
\begin{figure}[t]
	\begin{center}
		\includegraphics[width=0.99\linewidth]{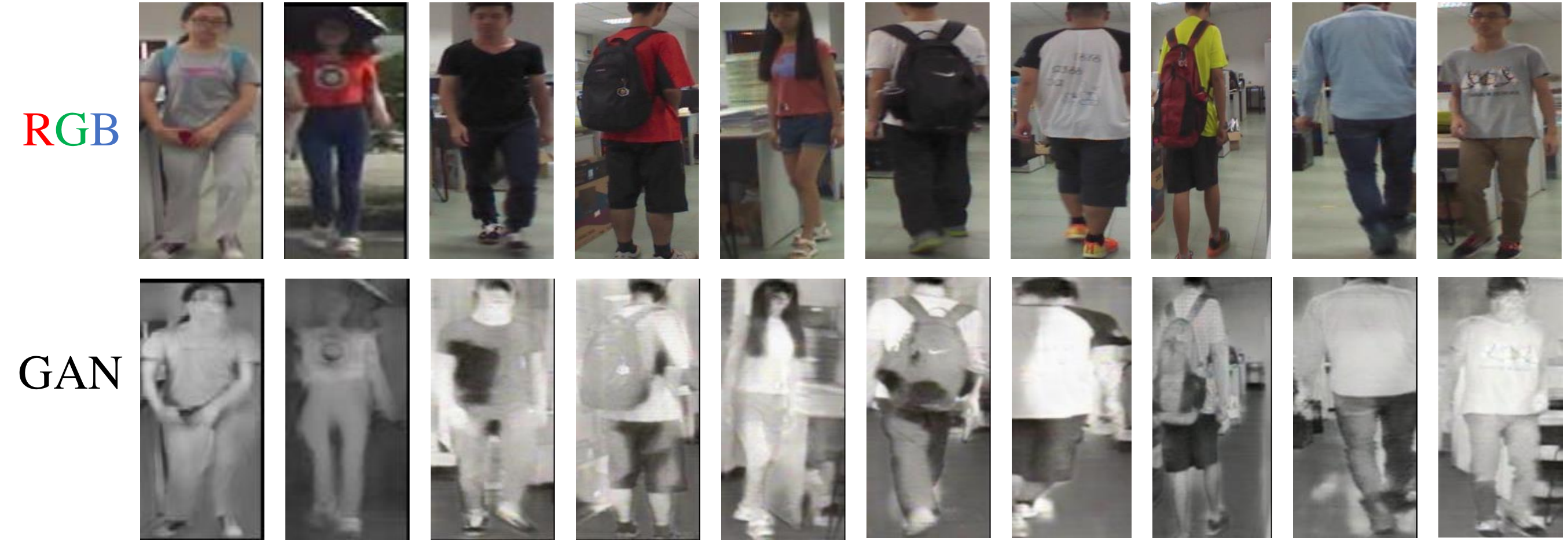}\vspace{-0.4cm}
	\end{center}
	\caption{Examples of translated images generated by CycleGAN on SYSU-MM01. The images in the first row are the original visible images and the second row are generated infrared images. Sampled images of the same column have the same identity.\vspace{-0.3cm}}
	\label{fig:smalltarget}
\end{figure}
\begin{table}[t]
	\centering
	\caption{Comparison of SFANet with different structures on the SYSU-MM01 dataset. Rank-1 matching accuracy ($\%$) and mAP ($\%$) are reported.}
	\label{Tab03}
	
	\begin{tabular}{l|cc|cc}	
		\hline
		& \multicolumn{2}{c}{All Search} & \multicolumn{2}{c}{Indoor Search} \\	
		\hline			
		{Shared layers}	&Rank-1  &mAP  &Rank-1  &mAP\\			
		\hline		
		Two-stream (None shared)  & 45.23& 42.43& 52.53& 58.68  \\
		One-stream (Fully shared)    &62.66&  56.22 &66.50  & 75.30\\
		MSTN \cite{33} (Four-layers shared)   &63.17&  56.29 &67.62  & 76.11\\
		Ours (Three-layers shared)  & \textbf{65.74}& \textbf{60.83} & \textbf{71.60}  &\textbf{80.05}  \\
		\hline		
	\end{tabular}
	
\end{table}
\begin{figure}[t]
	\begin{center}
		\includegraphics[width=0.99\linewidth]{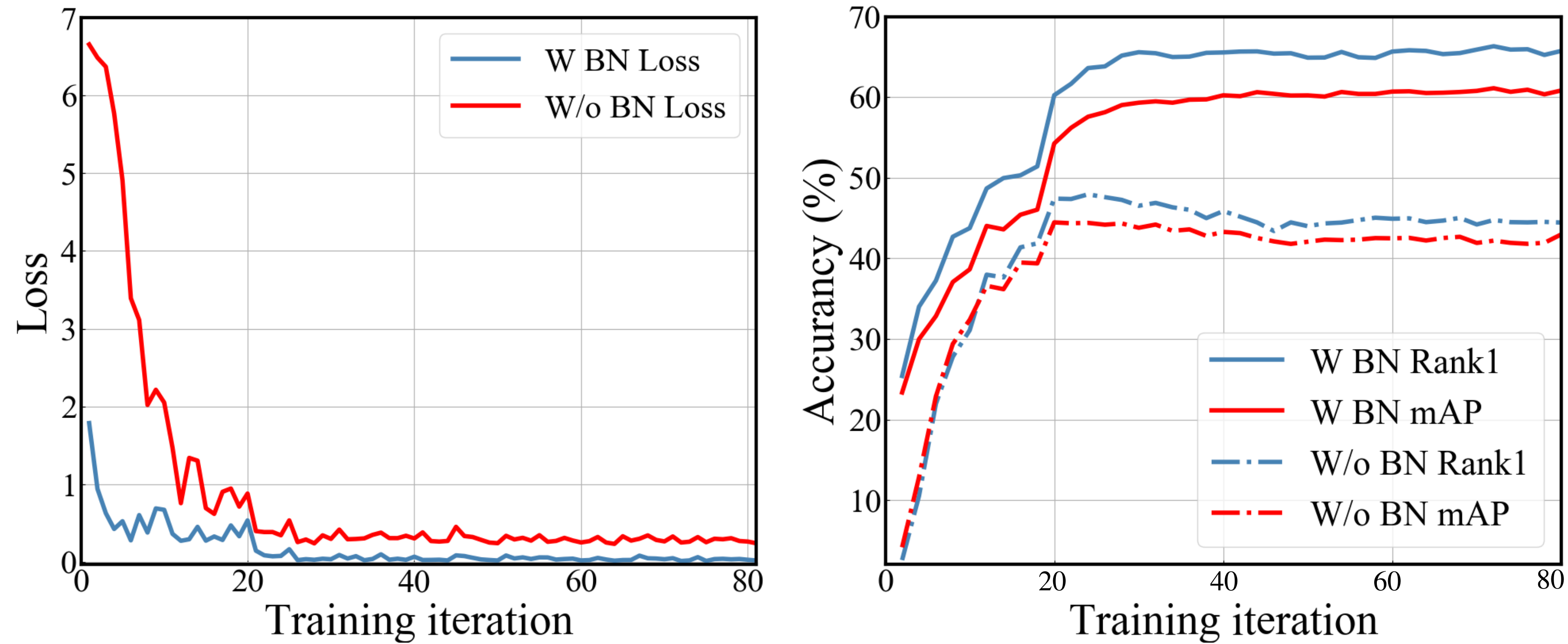}\vspace{-0.4cm}
	\end{center}
	\caption{Effectiveness of the BN layer for feature embedding on SYSU-MM01 dataset. We plot the loss curve (left) and report the rank-1 matching accuracy ($\%$) and mAP ($\%$) at different epochs. \vspace{-0.3cm}}
	\label{fig:smalltarget}
\end{figure}
\begin{figure*}[t]
	\begin{center}
		\includegraphics[width=0.99\linewidth]{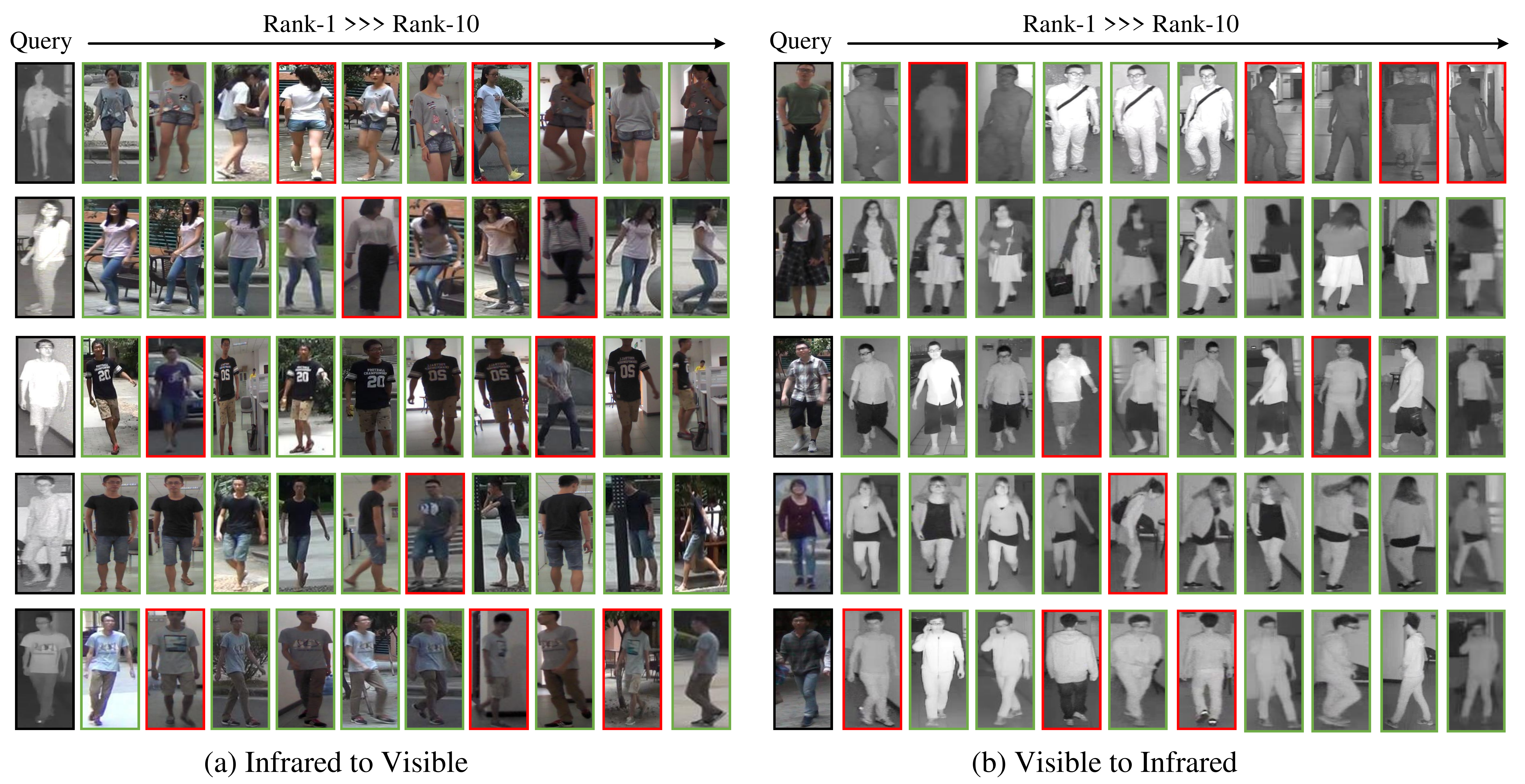}\vspace{-0.4cm}
	\end{center}
	\caption{Samples of top ten retrieved results with our proposed method on the test set of SYSU-MM01. The images with green borders belong to the same identity as the
		given query which has black border, red is opposite.\vspace{-0.3cm}}
	\label{fig:smalltarget}
\end{figure*}
\subsection{Analysis of the sharable dual-path information-preserving architecture.}
In VI-ReID community, almost all of algorithms are performed based on two types of network architectures, named one-stream network and two-stream network. In what follows we fully describe the characteristics of these network architectures and then analyse the improvement of our proposed sharable dual-path information-preserving architecture.

For widely-used two-stream structure, all the network parameters in the convolution blocks are optimized separately to capture modality-specific information and then use a fully connected layer to project these modality-specific feature vectors into a common space. However, in the authors'opinion, this design is overwhelmingly dependent on high-level sharable features in the final embedding layers and ignores the person spatial structure information which is crucial to descirbe a person. For another one-stream network where the shared structures are completely identical, it may implicitly learn  the modality-specific and shared information but fail to achieve low-level visual patterns for specific modality which contributes to miss some critical cues for similarity measure. In comparison, our sharable dual-path information-preserving structure that adopts two convolutional layers ($conv1$, $conv2$) to learn the modality-specific feature maps and the remaining three convolutional layers ($conv3$ $\sim$ $conv5$) are shared for two modalities to capture modality-shared information for feature embedding. This design effectively learns the 3D-shaped middle-level sharable feature vectors with perserving person spatial structure information to help the model learn discriminative feature representations for cross-modality matching.

We conduct two groups of VI-ReID experiments to apply our method to the standard two-stream and one-stream networks. In addition, we also uitilize the classic parameter sharing work named MSTN \cite{33} for comparsion. The results are shown in Table \Rmnum{7}. We observe that the sharable dual-path information-preserving structure has a superior performance compared to two-stream baseline network. This is because the partially shared convolutional blocks simultaneously capture the specific and shared feature representations rather than optimizing them independently. Moreover, benifiting from the parameter sharing parts, the person spatial structure information is preserved in the common space. Another observation is that the performance is also consistently improved by one-stream and MSTN network backbone. However, the improvement is not as siginificant as the sharable dual-path information-preserving architecture which improve the second best MSTN network by 2.57$\%$ of rank-1 accuacy and 4.54$\%$ of mAP. This experiment demonstrates that the modality-shared feature embedding netowrk with setting acceptable independent parameters of two blocks is more suitable than other choices for cross-modality modelling.


\subsection{Analysis of the Batch Normalization Operation.}
Previous work demonstrated that combining the identity loss and triplet loss can make the network learn better feature representations for Re-ID task. However, we find that the target of these two losses are always inconsistent in the embedding space. That means the gradient direction of two tasks may be inconsistent during iterations. The main reason is that $\mathcal{L}_{id}$ constructs several hyperplanes to separate the embedding space into different subspaces. So, cosine distance instead of Euclidean distance is always used in an identity loss to optimize network parameters. On the contrary, $\mathcal{L}_{bttr}$ based on a triplet distance optimization target is always computed by Euclidean distance to enhance intra-class tightness and inter-class separability in the Euclidean space. If we use both losses to optimize a feature space simultaneously, then one loss will be all the way down and the other will increase first and then decrease at one stage.

Previous literature point out the BN layer can overcome the overfitting and boosts the performance of IDE baseline. In authors'opinion, we consider that the BN layer can also smoothen the feature distribution in the embedding space. Assume that the feature $f_x$ before the BN layer that used to calculate $\mathcal{L}_{bttr}$ and the feature $f_y$ passing through the BN layer that used to calculate $\mathcal{L}_{id}$, the batch normalization layer with the extra scaling and shift paprameters helps recalibrate the channels of these embedding vectors. As a result, in the process of back propagation, $f_x$ not only can keep a compact distribution form but also acquires ID knowledge from ID loss and $f_y$ with batch normalization has clear decision surfaces due to the the weaker influence of $\mathcal{L}_{bttr}$. We report the results with and without BN layer in Fig. 10. It can be seen that adding a BN layer can apparently suppress the inconsistency, smoothen the loss curve and achieve a better optimization performance.
\begin{table}[t]
	\centering
	\caption{Comparsion with different triplet variants of BTTR loss on SYSU-MM01 dataset.}
	\label{Tab03}
	\begin{tabular}{l|cc|cc}	
		\hline
		& \multicolumn{2}{c}{All Search} & \multicolumn{2}{c}{Indoor Search} \\	
		\hline			
		{Strategy}	&Rank-1  &mAP  &Rank-1  &mAP\\			
		\hline		
		Baseline    &57.17&52.38&61.86&73.29  \\
		Triplet       &59.27&54.52&64.84&75.18 \\
		Triplet (Weighted) \cite{38}     &59.87&53.16 &65.74  & 74.41  \\
		Soft Margin \cite{37}       &61.23&55.17 &68.12  &76.82  \\
		\hline
		Tri-Hard \cite{37}  &\textbf{62.45}&\textbf{56.87} &\textbf{68.57}  &\textbf{ 77.95 } \\	
		\hline		
	\end{tabular}
\end{table}
\subsection{Comparsion with Other Triplet Variants.}
We conduct experiments to test the different variants of triplet training of bi-directional tri-constrained top-push ranking loss. In order to make this exploration tractable, we compare with other representative triplet variants, namely soft margin, weighted regularized triplet and lifted embedding losses. The baseline means the grayscale-infrared learning with only indentity loss. As shown in Table \Rmnum{8}, consistent improvements can be obtained by intehrating all proposed triplet re-formulations, which provides the relative distance optimization. It is nice to see that our BTTR loss with hard mining strategy achieves higher performance due to explicitly optimize hard samples for cross-modality and intra-modality relationships in different views, which ensures the discriminability of the learnt feature representation.

\subsection{Batch sampling strategy} In this paper, our batch sampling method is based on a $PK$ sampling strategy that randomly selects $P$ person identities and $K$ corresponding images, but we make some difference for optimizing the bi-directional tri-constrained top-push ranking (BTTR) loss. Specifically, at each training iteration, $\mathcal{N}$ person identities are randomly selected, where $\mathcal{N}$ is the batch size. Note that our network contains two input streams which come from different modalities respectively, we randomly sample one visible image and one infrared image to form the mini-batch. In this manner, a total of $2*\mathcal{N}$ images are obtained for network training. With the randomly sampling strategy, each image can be assigned to a non-overlap randomly identity so that all the possible assemblies will be traversed to get the global optimum.

\subsection{Visualization of results} We conduct visualization to show the top ten ranking results of the SYSU-MM01 test set, containing both infrared-to-visible and visible-to-infrared query settings. For each query setting, five query samples are randomly selected and their corresponding top ten retrieved cross-modality results are visualized in Fig. 11. The results show that our method has great robustness for modality and appearance discrepancies. The main reason is that the grayscale modality augmentation strategy perserves rich structural cues (e.g., bags or stripes) or conspicious part (e.g., logos). Even though there are some wrongly retrieved examples in the ranking list, our method can still retrieve the correct top-ranked images.

\section{Conclusion}
In this paper, we have presented a deep learning framework called SFANet for visible-infrared person re-identification. SFANet is formed based on the grayscale-spetrum feature augmentation strategy, which is captable of preserving the complete semantic information of RGB images and simultaneously alleviating the large modality discrepancy in the image space. Beyond the common two-stream feature extraction networks, we further extend SFANet with a parameter-sharing structure to capture the discriminative 3D-shaped spatial structure information of persons. In addition, we develop the feature-level constraints by proposing a dual-linear with batch normalization ID embedding method and a bi-directional tri-constrained top-push ranking (BTTR) loss, allowing to handle the modality difference in classifier-level and provide an effective metric measure to minimize the ambiguity between classes. Extensive experimental results on two standard benchmarks demonstrate that the proposed framework is robust enough to learn modality-invariant feature representations and outperforms state-of-the-art methods.
\section*{Acknowledgment}
This work is supported by the Nature Science Foundation of China (No. 61762023).





\begin{thebibliography}{1}
	\nocite{*}
	\bibliographystyle{IEEEbib}
	\bibitem{1}  Q. Leng, M. Ye, and Q. Tian, ``A survey of open-world person re-identification,'' \textit{IEEE TCSVT}, vol. 30, no. 4, pp. 1092-1108, 2019.
	
	\bibitem{2} M. Ye, J. Shen, G. Lin, T. Xiang, L. Shao, and S. C. H. Hoi, ``Deep learning for person re-identification: A survey and outlook,'' \textit{arXiv preprint arXiv:2001.04193}, 2020.
	
	\bibitem{3} M. Ye and J. Shen, ``Probabilistic Structural Latent Representation for Unsupervised Embedding,'' \textit{in CVPR}, 2020, pp. 5456-5465.
	
	\bibitem{4} W. Wang, J. Shen, X. Lu, S. C. Hoi, and H. Ling, ``Paying attention to video object pattern understanding,`` \textit{IEEE TPAMI}, pp. 1-1, 2020.
	
	\bibitem{5} A. Wu, W. Zheng, H. Yu, S. Gong, and J. Lai, ``RGB-infrared cross-modality person re-identification,'' \textit{in ICCV}, 2017, pp. 5380-5389.
	
	\bibitem{6} P. Dai, R. Ji, H. Wang, Q. Wu, and Y. Huang, ``Cross-modality person re-identification with generative adversarial training.'' \textit{in IJCAI}, 2018, pp. 677-683.
	
	\bibitem{7} G. Watson, A. Bhalerao, ``Person re-identification combining deep features and attribute detection.'' \textit{MTAP}, vol. 79, no. 9-10, pp. 6463-6481, 2020.
	
	\bibitem{8} Y. Tian, Q. Li, D. Wang, B. Wan, ``Robust joint learning network: improved deep representation learning for person re-identification.'' \textit{MATP}, vol. 78, no. 17, pp. 24187-24203, 2019.
	
	\bibitem{9} L. Zhang, F. Liu, D. Zhang, ``Adversarial View Confusion Feature Learning for Person Re-identification,'' \textit{IEEE TCSVT}, PP. 1-1, 2020. 
	
	\bibitem{10} G. Wang, T. Zhang, J. Cheng, ``RGB-Infrared Cross-Modality Person Re-Identification via Joint Pixel and Feature Alignment.'' \textit{in ICCV}, 2019, pp. 3623-3632.
	
	
	\bibitem{11} M. Ye, X. Lan, J. Li, and Pong C. Yuen, ``Hi-erarchical discriminative learning for visible thermal person re-identification,'' \textit{in AAAI}, 2018.
	
	\bibitem{12} M. Ye, Z. Wang, X. Lan, and Pong C. Yuen, ``Visible thermal person re-identification via dual-constrained top-ranking,'' \textit{in IJCAI}, 2018, pp. 1092-1099.
	
	\bibitem{13} P. Dai, R. Ji, H. Wang, ``Cross-Modality Person Re-Identification with Generative Adversarial Training.'' \textit{in IJCAI}, 2018, pp. 677–683.
	
	\bibitem{14} Z. Wang, Y. Zheng, Y. Chuang, ``Learning to reduce dual-level discrepancy for infrared-visible person re-identification,'' \textit{in CVPR}, 2019, pp. 618-626.
	
	\bibitem{15} X. Zhong, T. Lu, W. Huang, ``Visible-Infrared Person Re-Identification via Colorization-Based Siamese Generative Adversarial Network,'' \textit{in ICMR}, 2020, pp. 421-427.
	
	
	\bibitem{16} G. Wang, Y. Yang, T. Zhang, ``Cross-Modality Paired-Images Generation for RGB-Infrared Person Re-Identification,'' \textit{Neural Netw}, vol. 128, pp. 294-304, 2020.
	
	\bibitem{17} W. Zhang, X. He, W. Lu, H. Qiao and Y. Li, "Feature Aggregation With Reinforcement Learning for Video-Based Person Re-Identification," \textit{IEEE TNNLS}, vol. 30, no. 12, pp. 3847-3852, 2019.
	
	\bibitem{18} L. Wei, Z. Wei, Z. Jin, ``SIF: Self-Inspirited Feature Learning for Person Re-Identification,'' \textit{IEEE TIP}, vol. 29, pp. 4942-4951, 2020.
	
	\bibitem{19} Z. Wang, J. Jiang, Y. Wu, M. Ye, X. Bai and S. Satoh, ``Learning Sparse and Identity-Preserved Hidden Attributes for Person Re-Identification,'' \textit{IEEE TIP}, vol. 29, pp. 2013-2025, 2020.
	
	\bibitem{20} L. An, X. Chen, S. Yang and X. Li, "Person Re-identification by Multi-hypergraph Fusion," \textit{IEEE TNNLS}, vol. 28, no. 11, pp. 2763-2774, 2017.	
	
	\bibitem{21} F. Liu, and L. Zhang, ``View Confusion Feature Learning for Person Re-Identification,'' in ICCV, 2019, pp. 6638-6647.
	
	\bibitem{22} S. Liao and S. Li, ``Efficient PSD constrained asymmetric metric learning for person re-identification,'' \textit{in ICCV}, 2015, pp. 3685-3693.
	
	\bibitem{23} Z. Feng, J. Lai, and X. Xie, ``Learning modality-specific representations
	for visible-infrared person re-identification,'' \textit{IEEE TIP}, vol. 29, pp. 579–590, 2020.
	
	\bibitem{24} M. Ye, X. Lan, Z. Wang, and P. C. Yuen, ``Bi-directional center-constrained
	top-ranking for visible thermal person re-identification,'' \textit{IEEE TIFS}, vol. 15, pp.407-419, 2020. 
	
	\bibitem{25} D. Li, X. Wei, X. Hong and Y. Gong, ``Infrared-Visible Cross-Modal Person Re-Identification with an X Modality,'' \textit{in AAAI}, 2020, pp. 4610-4617.
	
	\bibitem{26} X. Wang, R. Girshick, A. Gupta, and K. He, ``Non-local neural networks,'' \textit{in CVPR}, 2018, pp. 7794-7803.
	
	\bibitem{27} D. Nguyen, H. Hong, K. Kim, K. Park, ``Person recognition system based on a combination of body images from visible light and thermal cameras,'' \textit{Sensors}, vol. 17, no. 3, pp. 605,  2017.
	
	\bibitem{28} L. Zheng, L. Shen, L. Tian, S. Wang, J. Wang, and Q. Tian, ``Scalable person re-identification: A benchmark,'' \textit{in ICCV}, 2015, pp. 1116-1124.
	
	\bibitem{29} K. He, X. Zhang, S. Ren, and J. Sun, ``Deep residual learning for image recognition,'' in CVPR, 2016, pp. 770-778.
	
	\bibitem{30} H. Luo, W. Jiang, Y. Gu, F. Liu, X. Liao, S. Lai, and J. Gu, ``A strong
	baseline and batch normalization neck for deep person re-identification,''
	\textit{arXiv preprint arXiv:1906.08332}, 2019.
	
	\bibitem{31} Lin, J.W., Li, H.: Hpiln: A feature learning framework for cross-modality person
	re-identification, \textit{IET Image Process}, vol. 13, no. 14, pp. 2897-2904, 2019.
	
	\bibitem{32} E. Basaran, and Mustafa E. Kamasak, ``An efficient framework for visible-
	infrared cross modality person re-identification,'' Signal Process Image Commun, vol. 87, pp. 115933, 2020. 
	
	\bibitem{33} M. Ye, X. Lan, Q. Leng and J. Shen, ``Cross-Modality Person Re-Identification via Modality-Aware Collaborative Ensemble Learning,'' \textit{IEEE TIP}, vol. 29, pp. 9387-9399, 2020.
	
	\bibitem{34} M. Ye, J. Shen, L. Shao, ``Dynamic Dual-Attentive Aggregation Learning for Visible-Infrared Person Re-Identification,'' \textit{in ECCV}, 2020.
	
	\bibitem{35} Y. Hao, N. Wang, J. Li, X. Gao, ``HSME: Hypersphere manifold embed-
	ding for visible thermal person re-identification,'' \textit{in AAAI}, 2019, pp. 8385-8392.
	
	\bibitem{36} M. Ye, X. Lan, Q. Leng, ``Modality-aware collaborative learning for
	visible thermal person re-identification,'' \textit{in ACM MM}, 2019, pp. 347-355.
	
	\bibitem{37} A. Hermans, L. Beyer, and B. Leibe, ``In defense of the triplet loss for
	person re-identification,'' \textit{arXiv preprint arXiv:1703.07737}, 2017.
	
	\bibitem{38} X. Wang, X. Han, W. Huang, D. Dong, and M. R. Scott, ``Multi-
	similarity loss with general pair weighting for deep metric learning,'' in CVPR, 2019, pp. 5022–5030.
	
	\bibitem{39} H. Song, Y. Xiang, S. Jegelka, and S. Savarese, ``Deep
	Metric Learning via Lifted Structured Feature Embedding,''
	\textit{In CVPR}, 2016, pp. 4004-4012.
	
	\bibitem{40} L. Wu, Y. Wang, L. Shao and M. Wang, "3-D PersonVLAD: Learning Deep Global Representations for Video-Based Person Reidentification," \textit{IEEE TNNLS}, vol. 30, no. 11, pp. 3347-3359, 2019.
	
	\bibitem{41} Z. Wang, J. Jiang, Y. Wu, M. Ye, X. Bai and S. Satoh, ``Learning Sparse and Identity-Preserved Hidden Attributes for Person Re-Identification,'' \textit{IEEE TIP}, vol. 29, pp. 2013-2025, 2020.
	
	\bibitem{42} X. Yang, P. Zhou and M. Wang, "Person Reidentification via Structural Deep Metric Learning," \textit{IEEE TNNLS}, vol. 30, no. 10, pp. 2987-2998, 2019.
	
	\bibitem{43} X. Yang, M. Wang and D. Tao, ``Person Re-Identification With Metric Learning Using Privileged Information,'' \textit{IEEE TIP}, vol. 27, no. 2, pp. 791-805, 2018.
	
	\bibitem{44} Y. Lu, Y. Wu, B. Liu, T. Zhang, and B. Li, ``Cross-Modality Person Re-Identification With Shared-Specific Feature Transfer.'' \textit{in CVPR}, 2020, pp. 13379-13389.
	
	\bibitem{45} H. Liu, H. Cheng, ``Enhancing the discriminative feature learning for
	visible-thermal cross-modality person re-identification,'' \textit{arXiv preprint arXiv:1907.09659}, 2019.
	
	\bibitem{46} X. Fan, H. Luo, C. Zhang, W. Jiang, ``Cross-Spectrum Dual-Subspace Pairing for RGB-Infrared Cross-Modality Person Re-Identification.'' \textit{ArXiv Preprint ArXiv:2003.00213}, 2020.
	
	\bibitem{47} H. Liu, X.Tan and X.Zhou, ``Parameter Sharing Exploration and Hetero-center Triplet Loss for Visible-Themal Person Re-identification,'' \textit{IEEE TMM}, 2020.
	
\end{thebibliography}
\end{document}